\documentclass{article}

\usepackage{microtype}
\usepackage{graphicx}
\usepackage{subcaption}
\usepackage{booktabs}

\usepackage{hyperref}
\usepackage{multirow}

\usepackage[accepted]{icml2026}

\usepackage{amsmath}
\usepackage{amssymb}
\usepackage{mathtools}
\usepackage{amsthm}
\usepackage{dsfont}
\usepackage{enumitem}
\usepackage{algorithm}
\usepackage{pifont}
\usepackage[table]{xcolor}

\usepackage[capitalize,noabbrev]{cleveref}

\theoremstyle{plain}

\theoremstyle{definition}

\theoremstyle{remark}

\usepackage[textsize=tiny]{todonotes}

\icmltitlerunning{LithoGRPO: Fast Inverse Lithography via GRPO Reinforced Flow Matching}

\begin{document}

\twocolumn[
  \icmltitle{LithoGRPO: Fast Inverse Lithography via GRPO Reinforced Flow Matching}

  \icmlsetsymbol{equal}{*}

  \makeatletter
  \newcounter{@affilhku}\setcounter{@affilhku}{1}
  \newcounter{@affilcambridge}\setcounter{@affilcambridge}{2}
  \newcounter{@affilcuhk}\setcounter{@affilcuhk}{3}
  \newcounter{@affilnjupt}\setcounter{@affilnjupt}{4}
  \newcounter{@affilhust}\setcounter{@affilhust}{5}
  \setcounter{@affiliationcounter}{5}
  \makeatother

  \begin{icmlauthorlist}
    \icmlauthor{Yao Lai}{hku,cambridge}
    \icmlauthor{Xuyuan Xiong}{hku}
    \icmlauthor{Zeyue Xue}{hku}
    \icmlauthor{Guojin Chen}{cuhk}
    \icmlauthor{Jing Wang}{njupt} \\
    \icmlauthor{Xihui Liu}{hku}
    \icmlauthor{Rui Zhang}{hust}
    \icmlauthor{Robert Mullins}{cambridge}
    \icmlauthor{Bei Yu}{cuhk}
    \icmlauthor{Ping Luo}{hku}
  \end{icmlauthorlist}

  \icmlaffiliation{hku}{The University of Hong Kong, Hong Kong}
  \icmlaffiliation{cambridge}{University of Cambridge, Cambridge, UK}
  \icmlaffiliation{cuhk}{The Chinese University of Hong Kong, Hong Kong}
  \icmlaffiliation{njupt}{Nanjing University of Posts and Telecommunications, Nanjing, China}
  \icmlaffiliation{hust}{School of Computer Science and Technology, Huazhong University of Science and Technology, Wuhan, China}

  \icmlcorrespondingauthor{Ping Luo}{pluo@cs.hku.hk}
  \icmlcorrespondingauthor{Bei Yu}{byu@cse.cuhk.edu.hk}
  \icmlcorrespondingauthor{Robert Mullins}{Robert.Mullins@cl.cam.ac.uk}

  \icmlkeywords{Machine Learning, ICML}

  \vskip 0.3in
]

\printAffiliationsAndNotice{}

\begin{abstract}
In semiconductor manufacturing, lithography projects circuit layouts onto silicon wafers through an optical mask.
As circuit features shrink below the wavelength of light, optical diffraction causes the printed patterns to deviate from their intended layouts.
Inverse Lithography Technology (ILT) addresses this challenge by generating optimized masks that enhance the fidelity of pattern transfer onto wafers.
While ILT resembles an image synthesis task, its reliance on explicit physical metrics for mask evaluation limits the applicability of existing generative models.
We introduce LithoGRPO, an ILT framework that integrates the flow‑matching paradigm with GRPO‑based reinforcement learning (RL) fine‑tuning, enabling efficient exploration of diverse masks for a given target layout.
Unlike purely generative or optimization‑based approaches,  RL in LithoGRPO exploits the explicitly defined, physics‑based reward function of ILT, enabling optimization under complex, process‑aware constraints.
To the best of our knowledge, this is the first framework that unifies flow matching and RL for mask optimization.
To improve RL sampling efficiency, we propose a fast shot-counting algorithm for manufacturability evaluation, achieving over 130$\times$ speedup while largely preserving the mask ranking of the traditional shot count metric.
Extensive experiments demonstrate that LithoGRPO achieves state‑of‑the‑art performance over both optimization‑based and learning‑based methods, while maintaining efficient mask generation.
Code is available at \href{https://github.com/laiyao1/LithoGRPO}{github.com/laiyao1/LithoGRPO}.

\end{abstract}

\section{Introduction}
\label{sec:intro}

\begin{figure}[!t]
  \centering
  \includegraphics[width=\columnwidth]{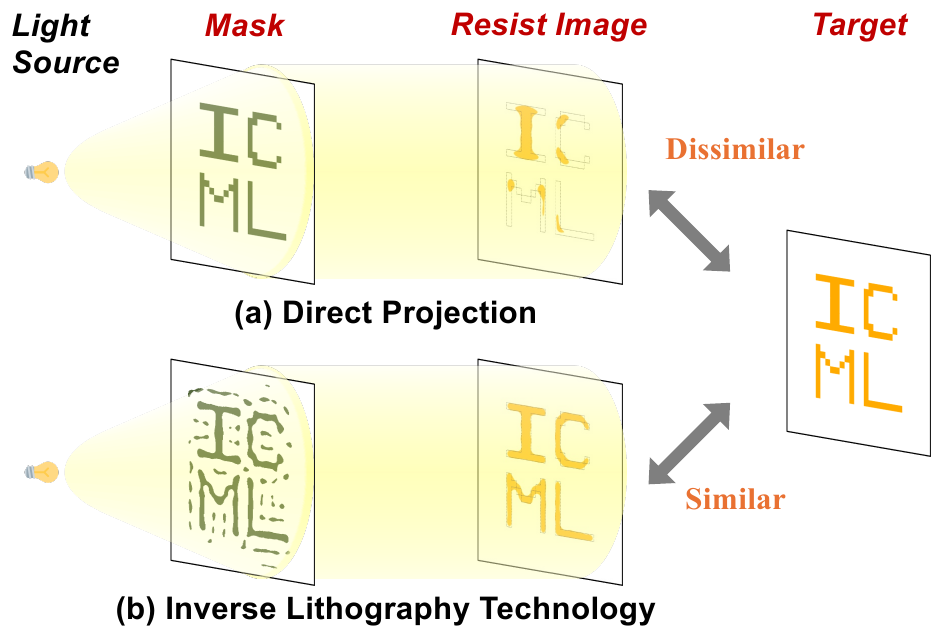}
\caption{\textbf{Inverse Lithography.}
(a) Direct projection suffers from pattern distortions in the resist image 
due to optical and process limitations.
(b) Inverse Lithography Technology (ILT) optimizes the mask to compensate 
for these distortions, accurately reproducing the target layout on the wafer.}
   \label{fig:demo}
\end{figure}

Lithography is a fundamental process in semiconductor manufacturing that prints circuit layouts onto silicon wafers~\cite{levinson2005principles, Erdmann2021OpticalAE}.
As illustrated in Fig.~\ref{fig:demo}, a light source projects the mask pattern onto a light‑sensitive photoresist layer covering the wafer.
After exposure and development, the remaining resist image defines the printed circuit layout.
With the continued scaling of technology nodes following Moore’s law, pattern features have become smaller than the exposure wavelength.
As a result, direct optical projection suffers from diffraction and optical aberrations, causing the printed patterns to deviate from the intended layouts, as shown in Fig.~\ref{fig:demo}(a).

\begin{figure*}[!th]
  \centering
  \begin{subfigure}[t]{0.186\textwidth}
    \centering
    \includegraphics[width=\textwidth]{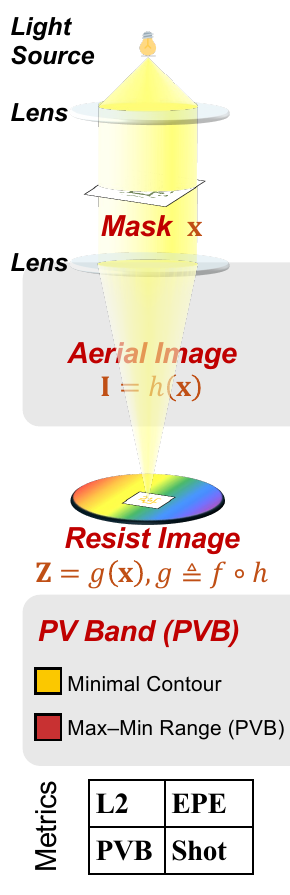}
    \label{fig:teaser_left}
  \end{subfigure}
  \hfill
  \begin{subfigure}[t]{0.799\textwidth}
    \centering
    \includegraphics[width=\textwidth]{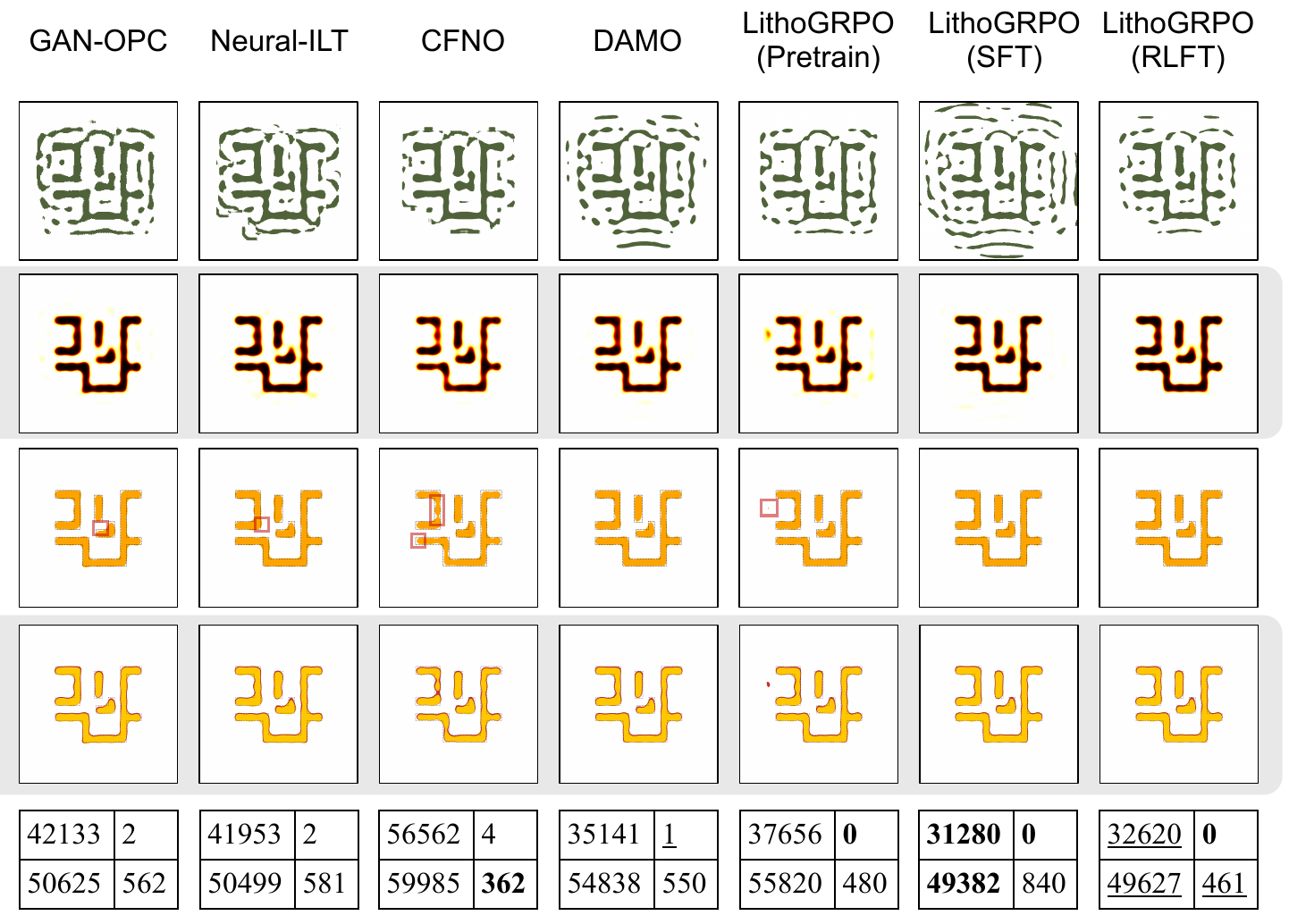}
    \label{fig:teaser_right}
  \end{subfigure}
\caption{\textbf{ILT results visualization and comparison.}
(Left) Illustration of the lithography. 
(Right) Results of different ILT methods for the same target layout. 
For the \textit{aerial image}, color intensity indicates the light exposure level.
For the \textit{resist image}, dashed contours denote the target layout, and yellow regions indicate the simulated resist image, with critical imaging differences highlighted by red boxes.
The \textit{PV Band} highlights process‑variation regions, where larger red areas indicate greater deviations.
\textbf{Best} and \underline{second‑best} results are marked in bold and with underlines.
\textbf{Better viewed at 400\% zoom.}}
  \label{fig:teaser}
\end{figure*}

Traditionally, optical proximity correction (OPC) mitigates lithographic distortions by locally adjusting mask geometries based on empirical rules or pre‑calibrated models.
Rule‑based OPC applies handcrafted correction heuristics but lacks global optimization and flexibility for complex patterns, whereas model‑based OPC employs imaging simulations for iterative edge refinement, achieving higher accuracy yet remaining limited to edge‑level corrections (i.e., local displacements of existing mask edges)~\cite{poonawala2007mask,pang2021inverse,yang2025advancements}.
Recent efforts to enhance OPC adaptability include RL‑OPC~\cite{liang2024rlopc}, which leverages reinforcement learning to directly optimize explicit performance metrics and improve correction capability while still operating at the edge level within the OPC framework.

To overcome the limited, edge‑based correction capability of OPC, Inverse Lithography Technology (ILT) formulates mask synthesis as a pixel‑level inverse imaging problem, enabling systematic, physically‑based correction of optical and process effects beyond the reach of OPC~\cite{yang2025advancements}, as shown in Fig.~\ref{fig:demo}(b).
Existing ILT methods generally fall into two categories: optimization‑based and learning‑based approaches.
Optimization‑based methods~\cite{gao2014mosaic, yu2022gpu, sun2023efficient, sun2025adaptive, jia2010machine} model the lithography imaging process as a differentiable system and optimize differentiable objectives through gradient descent. However, they cannot directly handle non‑differentiable objectives, and the iterative optimization process results in high computational cost.
Learning‑based methods~\cite{zhu2023l2o, yang2022generic, yang2022large, chen2020damo, jiang2020neural, jiang2021neural, wu2025tokman, wu2025lvmmo, jin2025unitho} aim to improve efficiency by training image‑to‑image models from paired masks and target layouts, but their training data typically originates from optimization‑based results, limiting the quality and generalization capability.
Furthermore, these methods still rely on differentiable losses, leaving non‑differentiable objectives unoptimized.
Among learning-based approaches, diffusion‑based variants~\cite{haoyu2025hifimo, haoxiang2025bbdm} achieve high‑fidelity synthesis but suffer from slow multi‑step sampling, constraining scalability for high-resolution ILT.
ILILT~\cite{yang2024ililt} uses a hybrid optimization–learning framework to improve end-to-end prediction fidelity, but it is computationally expensive and cannot optimize non-differentiable metrics.

\begin{figure*}[!t]
  \centering
  \includegraphics[width=\textwidth]{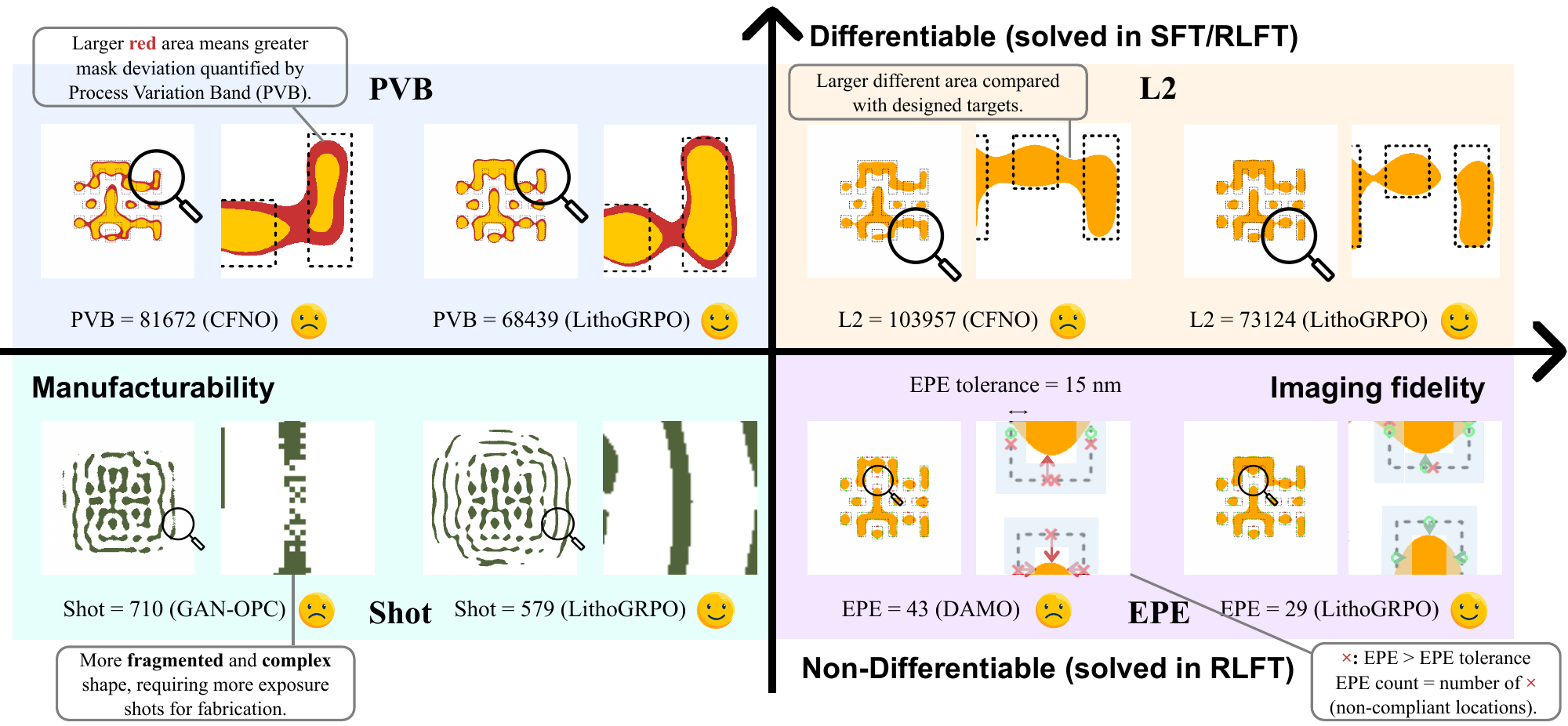}
\caption{\textbf{ILT evaluation metrics.}
$\mathrm{L2}$ and $\mathrm{EPE}$ assess imaging fidelity: 
$\mathrm{L2}$ measures pixel differences from the target, while $\mathrm{EPE}$ checks edge deviations within a threshold (Max EPE). 
$\mathrm{Shot}$ evaluates mask manufacturability, and $\mathrm{PVB}$ measures process‑variation robustness. 
$\mathrm{L2}$ and $\mathrm{PVB}$ are differentiable w.r.t.\ the mask (optimized in SFT/RLFT), 
whereas $\mathrm{EPE}$ and $\mathrm{Shot}$ are non‑differentiable and handled via RLFT. 
These metrics are interrelated and may conflict. 
For instance, improving imaging fidelity ($\mathrm{L2}$/$\mathrm{EPE}$) typically requires finer mask features, which increases the $\mathrm{Shot}$ count.
}
  \label{fig:metric}
\end{figure*}

To tackle these challenges, we propose LithoGRPO, a flow‑matching–based generative ILT framework optimized via GRPO‑driven reinforcement learning (RL), inspired by recent GRPO‑guided image synthesis methods~\cite{shao2024deepseekmath, guo2025deepseek, liu2025flow, xue2025dancegrpo, li2025mixgrpo}.
It performs reward-guided finetuning of generative models for mask synthesis, enabling unified optimization over differentiable and non-differentiable metrics. 
Such a setup is particularly suitable for lithography, where explicit and deterministic metrics naturally serve as well-defined rewards for reinforcement optimization. 
Unlike prior work, LithoGRPO models complete mask generation as a rectified flow~\cite{liu2022flow}, offering a deterministic one-step transport between noise and data that avoids the slow, multi-step denoising of diffusion models and improves sampling efficiency and stability. 
The training consists of three stages: (1) pretraining for mask-target alignment; (2) supervised finetuning optimizing differentiable metrics; and (3) RL finetuning jointly optimizing differentiable and non-differentiable metrics via SDE-based GRPO exploration. 
We further propose an ultra-fast shot-count evaluation based on minimal-overlap rectangular decomposition, reducing metric computation from about 1 min to 0.2 s. 
Overall, LithoGRPO unifies gradient‑based learning and reinforcement optimization under a rectified‑flow framework, jointly optimizing differentiable and non‑differentiable metrics for efficient and comprehensive ILT.

Our main contributions are summarized as follows:
\begin{itemize}
[itemsep=1pt, topsep=1pt, parsep=0pt, partopsep=0pt]

\item To the best of our knowledge, LithoGRPO is the \textit{first} to introduce flow matching into ILT by modeling mask generation as a rectified flow matching process conditioned on the target layout.

\item A GRPO‑based reinforcement learning fine‑tuning (RLFT) scheme tailored for ILT tasks with explicit reward definitions, enabling joint optimization of differentiable and non‑differentiable lithographic metrics under a unified framework.

\item An ultra-fast shot count algorithm speeds up mask evaluation by 130×--490× while largely preserving ranking, significantly improving RL sampling efficiency.

\item Extensive experiments demonstrate that LithoGRPO achieves state-of-the-art performance in imaging fidelity, manufacturability, and robustness, while maintaining high generation efficiency.
\end{itemize}

\section{Preliminary}
\label{sec:prelim}

\subsection{Inverse Lithography Technology}

As shown on the left side of Fig.~\ref{fig:teaser}, the lithography process projects a mask pattern $\mathbf{x}$ through an optical system onto a photoresist‑coated wafer, forming an aerial image $\mathbf{I}$ on the resist surface and a resist image $\mathbf{Z}$ on the wafer after development.

\textbf{Optical Imaging and Photoresist Modeling.}
Instead of building a full physical simulation, people usually adopt a compact differentiable forward model that links the mask pattern $\mathbf{x}$ to the printed resist image $\mathbf{Z}$ via the optical imaging and photoresist processes.
The aerial image $\mathbf{I}$ shown in Fig.~\ref{fig:teaser} can be described by Hopkins' diffraction model~\cite{Hopkins1953diffraction}:
\begin{equation}
\mathbf{I} = h(\mathbf{x})=\sum_{k=1}^{K} \mu_k \, \big| \, h_k \otimes \mathbf{x} \, \big|^2 ,
\label{eq:hopkins}
\end{equation}
where $\mathbf{x}$ denotes the mask pattern, $h_k$ is the $k$‑th coherent point‑spread function, and $\mu_k$ is its corresponding illumination weight.
The operator $\otimes$ denotes convolution, and $|\cdot|$ indicates the magnitude of a complex amplitude.
Intuitively, Eq.~(\ref{eq:hopkins}) expresses that the mask pattern is blurred by the optical system and integrated over different illumination angles to form the aerial image $\mathbf{I}$ on the wafer.

Modeling a negative photoresist, where exposed regions are retained, the aerial image $\mathbf{I}$ is converted into a developed resist pattern through thresholding:
\begin{equation}
\mathbf{Z} = \mathds{1}(\mathbf{I} > I_{\mathrm{th}}),
\label{eq:resist_binary}
\end{equation}
where $\mathds{1}(\cdot)$ denotes the indicator function and $I_{\mathrm{th}}$ is the exposure threshold.
To enable gradient-based optimization for differentiable metrics, this non-differentiable process can be replaced by an approximation:
\begin{equation}
\mathbf{Z} = f(\mathbf{I}) = \frac{1}{1 + \exp[-\alpha (\mathbf{I} - I_{\mathrm{th}} )]},
\label{eq:resist}
\end{equation}
where $\alpha$ controls the transition sharpness.
Together, Eqs.~(\ref{eq:hopkins})–(\ref{eq:resist}) form a differentiable forward model that maps a mask pattern $\mathbf{x}$ to its developed resist image $\mathbf{Z}$, expressed compactly as $\mathbf{Z} = g(\mathbf{x}) = f(h(\mathbf{x}))$, where $h(\cdot)$ and $f(\cdot)$ denote the optical imaging and photoresist, respectively.

\textbf{Inverse Lithography and Evaluation.}
Inverse lithography technology (ILT) aims to find the optimal mask $\mathbf{x}$ that yields a resist image $\mathbf{Z} = g(\mathbf{x})$ matching the desired target layout $\mathbf{T}$. 
Its performance can mainly be assessed by four representative metrics: $\mathrm{L2}$, $\mathrm{PVB}$, $\mathrm{EPE}$, and $\mathrm{Shot}$, as illustrated in Fig.~\ref{fig:metric}.

Among the differentiable objectives, the L2 loss is the most widely used metric to measure imaging fidelity. It is defined as a function of a mask $\mathbf{x}$ and a target layout $\mathbf{T}$, reflecting how well the simulated resist image $\mathbf{Z}=g(\mathbf{x})$ reproduces the target layout $\mathbf{T}$:
\begin{equation}
\mathrm{L2}(\mathbf{x}, \mathbf{T}) = \| g(\mathbf{x}) - \mathbf{T} \|_2^2 .
\label{eq:l2}
\end{equation}

To account for process variations such as focus and dose fluctuation, the process variation band (PVB) evaluates the stability of the printed results for a given mask $\mathbf{x}$ across different process corners. It measures the intensity range between the best and worst process conditions:
\begin{equation}
\mathrm{PVB}(\mathbf{x}) = \| g_{\max}(\mathbf{x}) - g_{\min}(\mathbf{x}) \|_2^2 ,
\label{eq:pvb}
\end{equation}
where $g_{\max}(\mathbf{x})$ and $g_{\min}(\mathbf{x})$ are the simulated images from the same $\mathbf{x}$ under the best and worst process conditions, respectively. 
A larger PVB indicates higher sensitivity to process variations and thus lower robustness.

Also, two non‑differentiable metrics are critical for evaluating the local geometric fidelity and mask manufacturability. 
Edge Placement Error (EPE) measures the local deviation between the printed contour and the target edge. It is typically evaluated at discrete sampling points along the target boundary by counting the number of sites where the deviation exceeds a specified tolerance (Max EPE), with smaller EPE values indicating higher pattern fidelity.
The shot count (Shot) assesses mask manufacturability. It is the minimum number of rectangular shots required to cover the ILT pattern during e‑beam mask writing, directly impacting fabrication time and cost.
These metrics often exhibit inherent trade-offs: pursuing higher imaging fidelity (e.g., lower L2 / EPE) typically requires more complex mask geometries, which in turn increases mask writing complexity and degrades manufacturability,  leading to a larger Shot count.

\subsection{Flow Matching}

Flow-based generative modeling~\cite{lipman2024flow} has recently shown remarkable success in image and video synthesis, with models like Stable Diffusion 3~\cite{esser2024scaling} and FLUX~\cite{flux2024}.
These models operate by learning a continuous transformation that maps a simple prior distribution $p_0$ (e.g., a standard normal distribution $\mathcal{N}(\mathbf{0}, \mathbf{I})$) to a complex target data distribution $p_1$. This transformation is defined by an ordinary differential equation (ODE):
\begin{equation}
\frac{d \mathbf{x}_t}{d t} = \mathbf{v}_{\theta}(\mathbf{x}_t, t), \quad \mathbf{x}_0 \sim p_0,
\label{eq:flow_ode}
\end{equation}
where $\mathbf{v}_{\theta}(\mathbf{x}, t)$ is a neural network parameterized by $\theta$ that represents a time-dependent velocity field. Generation of new samples is accomplished by integrating this ODE from $t=0$ to $t=1$ using numerical solvers. 
For instance, the Euler method provides a first-order approximation:
\begin{equation}
\mathbf{x}_{t+\Delta t} = \mathbf{x}_t + \Delta t*\mathbf{v}_{\theta}(\mathbf{x}_t, t),
\label{eq:euler}
\end{equation}
which discretely simulates the probability path, or flow, that transports the prior distribution $p_0$ to the target distribution $p_1$. In contrast to diffusion models, this formulation establishes a more direct theoretical connection between the latent space and the data manifold, often enabling significantly faster sampling with fewer function evaluations.

Flow matching~\cite{lipman2023flow} introduces a general and simulation-free training paradigm for learning such ODEs. It trains the neural velocity field $\mathbf{v}_\theta$ by regressing it onto a predefined target vector field. A prominent variant, Rectified flow~\cite{liu2022flow}, simplifies the underlying dynamics by constructing linear interpolation paths between pairs of samples $(\mathbf{x}_0, \mathbf{x}_1)$ drawn from the prior and target distributions, respectively. The trajectory and its corresponding target velocity field are defined as:
\begin{equation}
\mathbf{x}_t = (1-t)\mathbf{x}_0 + t\mathbf{x}_1, \quad \mathbf{v}(\mathbf{x}_t, t) = \mathbf{x}_1 - \mathbf{x}_0.
\label{eq:rectified_traj}
\end{equation}
The training objective is thus formulated as a simple regression loss:
\begin{equation}
\begin{aligned}
\mathcal{L}_{\mathrm{flow}} = \mathbb{E}_{t \sim \mathcal{U}(0,1), \mathbf{x}_0 \sim p_0, \mathbf{x}_1 \sim p_1} \left[ \lVert \mathbf{v}_{\theta}(\mathbf{x}_t, t) - \mathbf{v}(\mathbf{x}_t, t)\rVert_2^2 \right].
\end{aligned}
\label{eq:rf}
\end{equation}
where the target velocity $\mathbf{x}_1 - \mathbf{x}_0$ remains constant along each individual linear trajectory. This rectification process yields substantially straighter and smoother flows, which are inherently easier to integrate and exhibit improved numerical stability during sampling.

\section{Method}

\begin{figure}[!ht]
    \centering
    \includegraphics[width=\columnwidth]{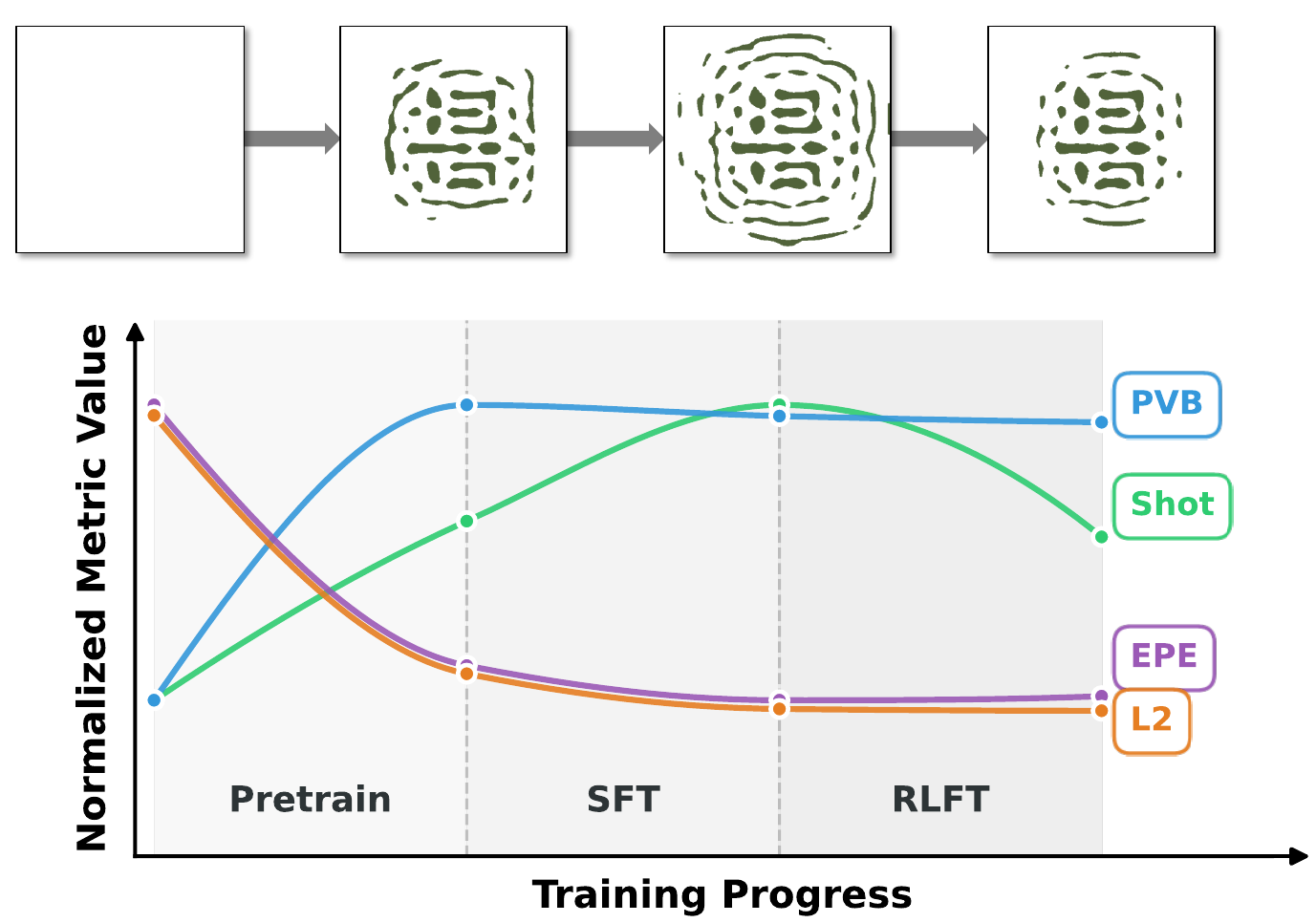}
    \caption{\textbf{Training dynamics of key metrics.} $\mathrm{L2}$ and $\mathrm{EPE}$ decrease during pre‑training and SFT, while $\mathrm{Shot}$ rises; the RLFT stage subsequently reduces $\mathrm{Shot}$ without degrading other metrics, highlighting the benefit of the three‑stage strategy.
    Curves are fitted from stage-end metric values.
    }
    \label{fig:training_metrics}
\end{figure}

\textbf{Overview.}
LithoGRPO is trained in three stages: pretraining, supervised fine-tuning (SFT), and reinforcement-learning fine-tuning (RLFT), as shown in Fig.~\ref{fig:training_metrics}.
Pretraining learns an initial ILT generator from a limited mask dataset. 
SFT adds differentiable lithographic metrics to the loss to improve pattern fidelity, at the cost of increased shot count due to metric trade-offs. 
RLFT then optimizes all metrics via a single reward, reducing shot count while largely preserving the other metrics near saturation.

\textbf{Pretraining and SFT.}
Following the two‑stage training framework in previous works~\cite{zheng2023lithobench}, we replace the original GAN‑based or pixel‑space generation modeling approach with rectified flow.

During pretraining, we use dataset‑provided pairs $(\mathbf{T}, \mathbf{x}_1)$,
where $\mathbf{T}$ denotes the target layout and $\mathbf{x}_1$ is the corresponding mask pattern.
For each pair, Gaussian noise $\mathbf{x}_0$ is sampled, and the model is trained with the standard rectified‑flow mean‑squared‑error objective to predict the velocity field $\mathbf{v}_\theta$ from $(\mathbf{x}_0, \mathbf{x}_1)$ conditioned on $\mathbf{T}$, as shown in Eq.~\ref{eq:rf}. 
In practice, the target layout $\mathbf{T}$ is concatenated with the noisy input $\mathbf{x}_t$ along the channel dimension.

In the supervised fine-tuning (SFT) stage, we incorporate differentiable, task-specific metrics into the optimization. 
At each training step, a random time $t \sim \mathcal{U}(0,1)$ is sampled, and the model predicts the velocity $\mathbf{v}_{\theta}(\mathbf{x}_t, t; \mathbf{T})$ from the interpolated state $\mathbf{x}_t$ conditioned on the target layout $\mathbf{T}$. 
The predicted velocity is then projected to the end of the trajectory to estimate the final mask:
\begin{equation}
\mathbf{x}_1 = \mathbf{x}_t + (1-t)\mathbf{v}_{\theta}(\mathbf{x}_t, t; \mathbf{T}),
\label{eq:prediction}
\end{equation}
where $\mathbf{x}_1$ denotes the predicted mask.
While the flow-matching loss $\mathcal{L}_{\mathrm{flow}}$ in Eq.~\ref{eq:rf} provides supervision at the intermediate time $t$, 
the differentiable metrics ($\mathrm{L2}$/$\mathrm{PVB}$) are computed on $\mathbf{x}_1$. 
The $\mathrm{L2}$ metric is computed between the target layout $\mathbf{T}$ and the resist image $\mathbf{Z}$ (Eq.~\ref{eq:l2}),
where $\mathbf{Z}=g(\mathbf{x}_1)$ is obtained via the differentiable lithography simulator. 
The $\mathrm{PVB}$ metric is similarly evaluated based on $\mathbf{x}_1$ (Eq.~\ref{eq:pvb}). 
The total SFT loss is a weighted combination:
\begin{equation}
\mathcal{L}_{\mathrm{sft}}
= \lambda_{\mathrm{flow}}\mathcal{L}_{\mathrm{flow}}
+ \lambda_{\mathrm{L2}}\mathrm{L2}(\mathbf{x}_1, \mathbf{T})
+ \lambda_{\mathrm{PVB}}\mathrm{PVB}(\mathbf{x}_1).
\label{eq:sft}
\end{equation}

\begin{figure}[!ht]
\centering
\includegraphics[width=\columnwidth]{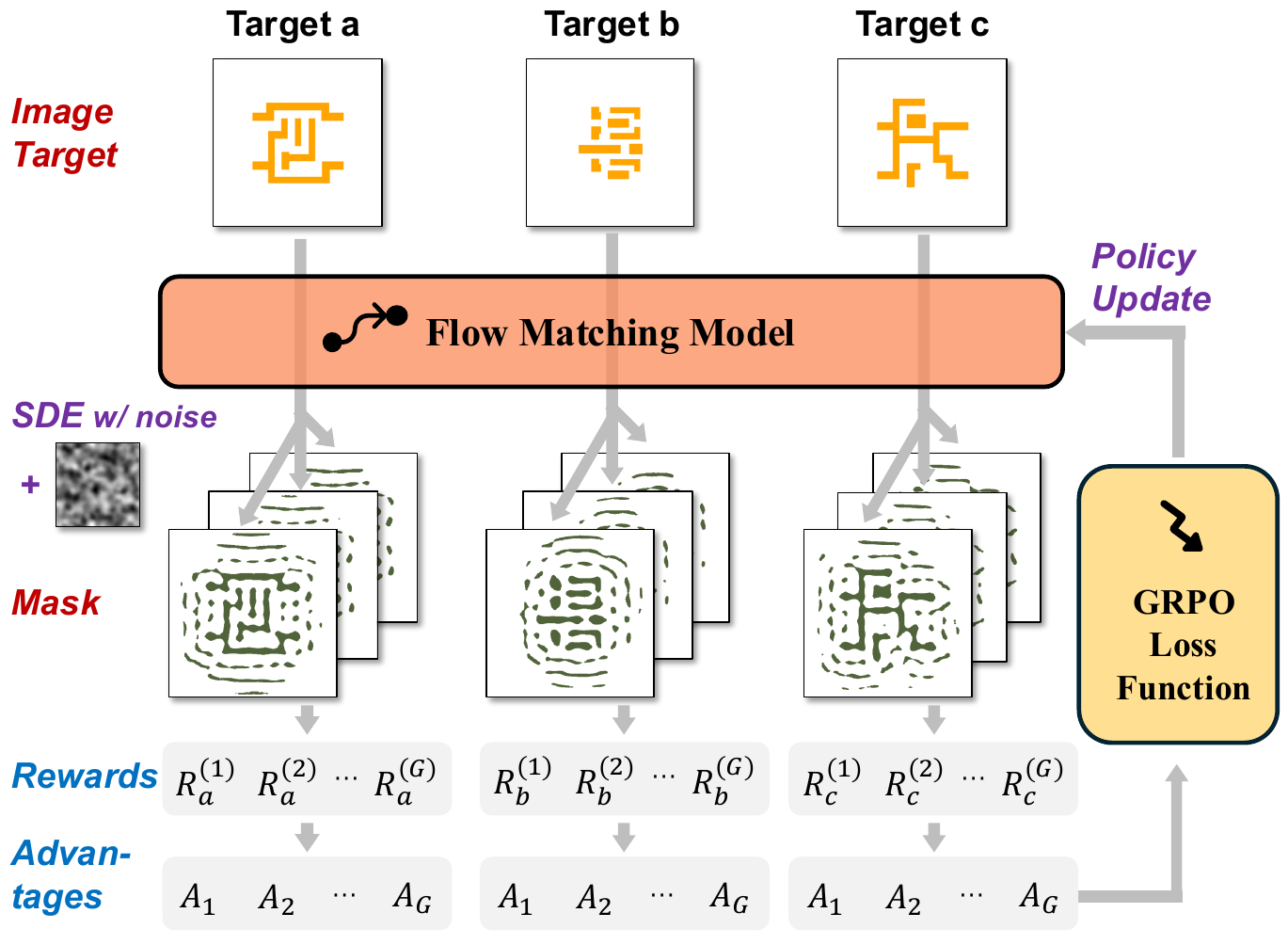}
\caption{\textbf{Overview of the RLFT stage in LithoGRPO.}
For each target layout, the flow matching model generates multiple mask 
samples by simulating SDE dynamics conditioned on the target and noise.
Each group of generated masks yields corresponding rewards 
$\{R^{(1)}, \ldots, R^{(G)}\}$, which are normalized to compute 
group-level advantages $\{A^{(i)}\}$.
These advantages are then used to update the policy via the GRPO loss.}
\label{fig:grpo_framework}
\end{figure}

\textbf{GRPO-based RLFT.}
In the third stage, we employ GRPO-based RL finetuning (RLFT) to jointly optimize all four metrics, including the non-differentiable ones ignored in earlier stages.
To achieve this, we draw inspiration from recent GRPO-based generative methods~\cite{xue2025dancegrpo,liu2025flow}. 
We reformulate the model's deterministic Ordinary Differential Equation (ODE) into an equivalent Stochastic Differential Equation (SDE). 
This SDE is constructed to preserve the marginal distributions of the original ODE at all time steps, while its inherent stochasticity enables the generation of multiple mask candidates for each target layout during RL optimization, as illustrated in Fig.~\ref{fig:grpo_framework}. 
Specifically, for a given target layout $\mathbf{T}$, the stochastic flow model samples a group of $G$ candidate mask patterns $\{\mathbf{x}_1^{(i)}\}_{i=1}^{G}$, for which we compute corresponding rewards $R(\mathbf{x}_1^{(i)}, \mathbf{T})$ using the all four performance metrics.
To reduce the variance of the policy gradient, we calculate the advantage for each mask candidate $\mathbf{x}_1^{(i)}$ by normalizing its reward against the statistics of the entire group:
\begin{equation}
A_i = \frac{R(\mathbf{x}_1^{(i)}, \mathbf{T}) - \operatorname{mean}(\{R(\mathbf{x}_1^{(i)}, \mathbf{T})\}_{i=1}^G)}{\operatorname{std}(\{R(\mathbf{x}_1^{(i)}, \mathbf{T})\}_{i=1}^G) + \varepsilon}.
\label{eq:advantage}
\end{equation}
This advantage $A_i$ then weights the score function of the corresponding trajectory, guiding the model to favor generations that yield higher rewards.

The reward signal for a generated mask $\mathbf{x}_1^{(i)}$ given a target $\mathbf{T}$ is defined as the negative sum of normalized metric scores. The metric set is $\mathcal{K} = \{\mathrm{L2}, \mathrm{PVB}, \mathrm{EPE}, \mathrm{Shot}\}$. 
Each metric $k$ is normalized by its baseline value $k_0$, obtained from the model at the end of the SFT stage.
As we adopt uniform weights, the reward is:
\begin{equation}
R(\mathbf{x}_1^{(i)}, \mathbf{T}) = - \sum_{k \in \mathcal{K}} k(\mathbf{x}_1^{(i)}, \mathbf{T})/k_0.
\label{eq:reward}
\end{equation}
The $\mathrm{Shot}$ and $\mathrm{PVB}$ metrics depend only on the generated mask $\mathbf{x}_1^{(i)}$, whereas $\mathrm{L2}$ and $\mathrm{EPE}$ additionally depend on the target layout $\mathbf{T}$, as introduced in the preliminaries.
For computational efficiency, we use a fast shot count approximation ($\mathrm{Shot}$(fast)) during RL training; the traditional shot count is used only for final evaluation.

We treat the generative model as a policy $\pi_\theta(\mathbf{x}\!\mid\!\mathbf{T})$, 
representing the conditional probability of generating layout $\mathbf{x}$ given a target $\mathbf{T}$, 
and optimize its parameters $\theta$ by minimizing the GRPO loss, defined as an expectation over the dataset of target layouts $\mathcal{D}$:
\begin{equation} \label{eq:grpo_loss}
\begin{gathered}
    \mathcal{L}_{\mathrm{grpo}}(\theta)
    = \mathbb{E}_{\mathbf{T} \sim \mathcal{D}} \Big[ f(\theta; \mathbf{T}) \Big], \quad \text{where} \\[1ex]
    f(\theta; \mathbf{T})
    = - \sum_{i=1}^{G}
      \min \!\Big( r_i(\theta) A_i,\,
      \operatorname{clip}\!\big(r_i(\theta), 1-\varepsilon, 1+\varepsilon\big) A_i \Big), \\[1ex]
    r_i(\theta)
    = \frac{\pi_\theta(\mathbf{x}_1^{(i)} | \mathbf{T})}
           {\pi_{\theta_{\text{old}}}(\mathbf{x}_1^{(i)} | \mathbf{T})}.
\end{gathered}
\end{equation}
Here $A_i$ is the advantage defined in Eq.~\ref{eq:advantage},
$\varepsilon$ is the clipping threshold, and
$\pi_{\theta_{\text{old}}}$ denotes the policy from the previous iteration used to compute the importance ratio.

\textbf{SDE-based Sampling in GRPO.}
To overcome the limited exploration inherent in the deterministic ODE-based sampling of our flow model, we introduce a stochastic generation process. Inspired by FlowGRPO~\cite{liu2025flow}, we reformulate the generation ODE into a Stochastic Differential Equation (SDE). This SDE is carefully designed to ensure its marginal distribution matches that of the original flow model at each timestep. By doing so, our sampler can explore a diverse range of candidates for GRPO optimization without deviating from the learned data manifold, thus preventing policy stagnation in local optima.

\begin{figure}[htbp]
    \centering
    \includegraphics[width=\columnwidth]{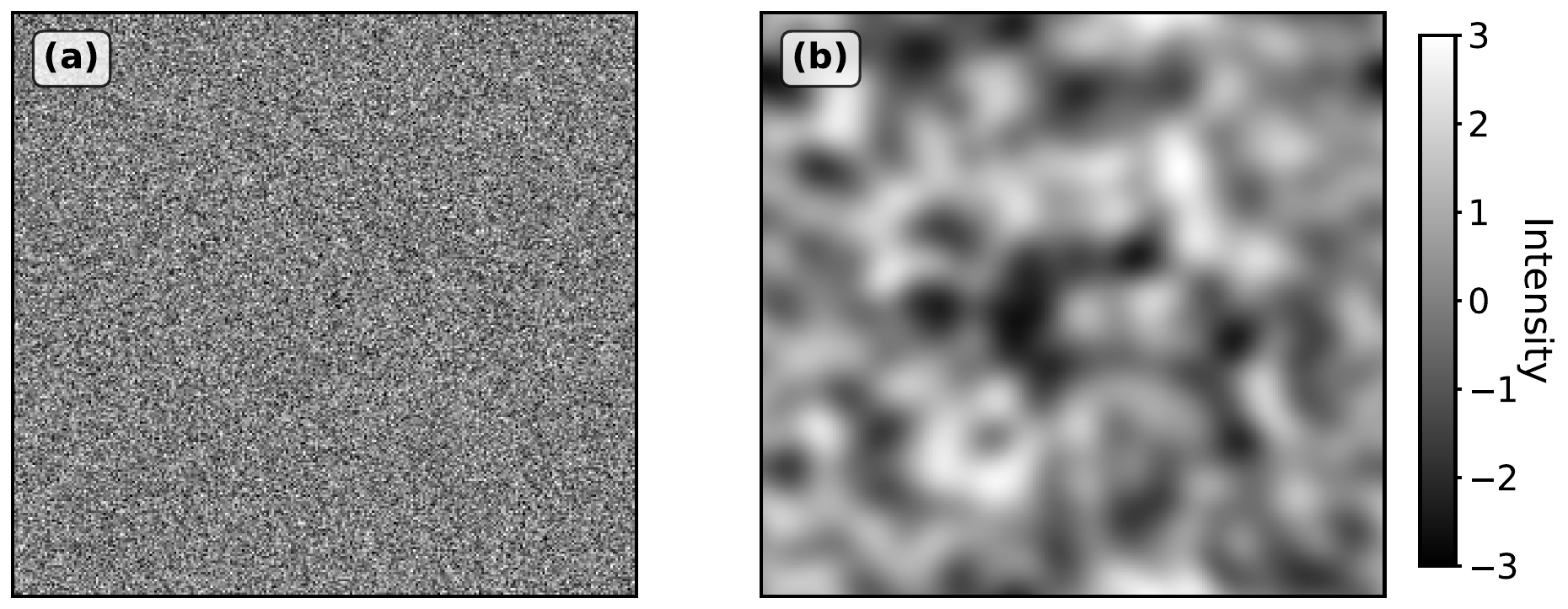}
\caption{\textbf{Comparison of noise types.}
(a) Gaussian white noise: high-frequency and spatially uncorrelated, 
leading to fragmented masks with high shot counts.
(b) Colored Gaussian noise: low-frequency and spatially correlated, 
preserving the spatial structure of lithography masks.}
    \label{fig:noise_comparison}
\end{figure}

Specifically, the reverse-time SDE is discretized using the Euler–Maruyama scheme~\cite{kloeden1977numerical}, 
yielding the following update rule for generating a sample $\mathbf{x}_{t+\Delta t}$ from the previous state $\mathbf{x}_t$:
\begin{equation}
\label{eq:sde_update_rule_split}
\begin{split}
\mathbf{x}_{t+\Delta t} = \mathbf{x}_t 
&+ \Big[ \mathbf{v}_{\theta}(\mathbf{x}_t, t) 
   + \frac{\sigma_t^2}{2t}\big( \mathbf{x}_t + (1-t)\mathbf{v}_{\theta}(\mathbf{x}_t, t) \big) \Big]\Delta t \\
&+ \sigma_t \sqrt{\Delta t}\,\boldsymbol{\varepsilon}.
\end{split}
\end{equation}
Here, the stochastic term $\boldsymbol{\varepsilon}$ is realized as low-frequency colored noise, 
with scale $\sigma_t = a\sqrt{(1-t)/t}$ where $a$ is a scalar hyperparameter.
Unlike the conventional setting of injecting Gaussian white noise 
($\boldsymbol{\varepsilon} \sim \mathcal{N}(0,\mathbf{I})$), 
uncorrelated noise introduces high-frequency perturbations in the mask domain, 
yielding fragmented regions that degrade manufacturability and increase shot count.
To alleviate this, colored noise is obtained by low-pass filtering white noise in the Fourier domain, 
producing spatially correlated perturbations that preserve mask topology and suppress spurious fragments. 
This design is crucial for ILT mask generation, where smooth and continuous features are essential. 
Fig.~\ref{fig:noise_comparison} qualitatively compares (a) unstructured white noise and (b) colored noise with correlated patterns.
For GRPO, we approximate each transition with an independent Gaussian for log-probability computation,
$\mathbf{x}_{t+\Delta t} \sim \mathcal{N}(\boldsymbol{\mu}_t, \sigma_t^2 \Delta t\,\mathbf{I})$,
where $\boldsymbol{\mu}_t$ denotes the deterministic component in Eq.~\ref{eq:sde_update_rule_split}.
Details are in Appendix~\ref{sec:log_prob_details}.

\begin{table*}[!th]
  \centering
\caption{\textbf{Performance comparison on ILT for Metal and Via layers.}
\textbf{Best} and \underline{second-best} are in bold and underlined.
LithoGRPO (RLFT) results are mean $\pm$ std over 4 independent trainings with different random seeds.
Avg. Rank is computed over 4 datasets $\times$ 4 metrics + Time (17 columns), with ties assigned the average of the occupied ranks.
}
  \label{tab:main_results_avg_time}
  \resizebox{\textwidth}{!}{%
  \begin{tabular}{l|rrrr|rrrr|rrrr|rrrr|c|c}
    \toprule
     & \multicolumn{8}{c|}{\textbf{Metal Layer}} & \multicolumn{8}{c|}{\textbf{Via Layer}} & \multirow{3}{*}{\textbf{\begin{tabular}[c]{@{}c@{}}Avg.\\Time\\(s)$\downarrow$\end{tabular}}} & \multirow{3}{*}{\textbf{\begin{tabular}[c]{@{}c@{}}Avg.\\Rank$\downarrow$\end{tabular}}} \\
    \cmidrule(lr){2-9} \cmidrule(lr){10-17}
    \textbf{Method} & \multicolumn{4}{c|}{MetalSet} & \multicolumn{4}{c|}{StdMetal (in-distribution)} & \multicolumn{4}{c|}{ViaSet} & \multicolumn{4}{c|}{StdContact (OOD)} & & \\
    \cmidrule(lr){2-5} \cmidrule(lr){6-9} \cmidrule(lr){10-13} \cmidrule(lr){14-17}
    & L2$\downarrow$ & PVB$\downarrow$ & EPE$\downarrow$ & Shot$\downarrow$ & L2$\downarrow$ & PVB$\downarrow$ & EPE$\downarrow$ & Shot$\downarrow$ & L2$\downarrow$ & PVB$\downarrow$ & EPE$\downarrow$ & Shot$\downarrow$ & L2$\downarrow$ & PVB$\downarrow$ & EPE$\downarrow$ & Shot$\downarrow$ & & \\
    \midrule
    \rowcolor{gray!15}
    \multicolumn{19}{l}{\textit{Optimization Approaches}} \\
    \midrule
    MOSAIC & 35860 & 48080 & 6.6 & 361 & 21733 & 27222 & 3.2 & \underline{229} & - & - & - & - & - & - & - & - & 0.940 & 9.8 \\
    LevelSet & 34712 & 50113 & 5.4 & \textbf{263} & 21526 & 27769 & 3.5 & \textbf{145} & 9632 & 10197 & 4.7 & \textbf{64} & 50770 & 32134 & 34.7 & \underline{275} & 2.290 & 6.9 \\
    MultiLevel & \textbf{27893} & 41372 & \textbf{2.5} & 1250 & \textbf{13203} & 23755 & \textbf{0.1} & 1240 & \textbf{4268} & 8370 & \textbf{0.0} & 1434 & 20497 & 43405 & 0.4 & 1473 & 1.030 & 5.6 \\
    \midrule
    \rowcolor{gray!15}
    \multicolumn{19}{l}{\textit{Learning Approaches}} \\
    \midrule
    GAN-OPC & 43414 & 41290 & 8.7 & 574 & 25929 & 23715 & 4.6 & 457 & 14767 & \textbf{6686} & 8.3 & \underline{166} & 81378 & \textbf{4931} & 73.2 & 276 & \textbf{0.010} & 7.4 \\
    Neural-ILT & 36670 & 42666 & 7.3 & 476 & 20045 & 23548 & 2.4 & 373 & 12723 & 8537 & 6.2 & 263 & 25422 & 41537 & 3.2 & \textbf{265} & \underline{0.025} & 6.5 \\
    DAMO & 32579 & 41173 & 5.4 & 523 & 16120 & 23796 & \underline{0.2} & 418 & 5081 & 9962 & \textbf{0.0} & 176 & 50445 & 35673 & 26.7 & 458 & 0.028 & 5.7 \\
    DOINN & 36409 & 41929 & 7.4 & 519 & 25913 & 25749 & 4.5 & 313 & 4382 & 7836 & \textbf{0.0} & 968 & 72058 & 17968 & 55.8 & 493 & 0.107 & 7.3 \\
    CFNO & 47814 & 46131 & 12.5 & \underline{302} & 26809 & 26814 & 4.2 & \underline{232} & 8949 & 9890 & \underline{0.1} & 184 & 70740 & 17950 & 55.1 & 396 & 0.040 & 7.4 \\
     \midrule
     \rowcolor{gray!15}
    \multicolumn{19}{l}{\textit{Mixed Approach}} \\
    \midrule
    ILILT & 30353 & 45353 & 3.2 & 433 & \underline{14596} & 24969 & \textbf{0.1} & 368 & 4666 & 10065 & \textbf{0.0} & 238 & 38957 & 43869 & 7.1 & 510 & 0.441 & 5.9 \\
    \midrule
    \rowcolor{gray!15}
    \multicolumn{19}{l}{\textbf{\textit{Our Approach}}} \\
    \midrule
    LithoGRPO (Pretrain) & 32824 & 42320 & 5.4 & 487 & 17362 & 23837 & 0.4 & 386 & 11595 & 8776 & 3.4 & 170 & 77289 & \underline{15445} & 63.9 & 377 & 0.104 & 6.6 \\
    LithoGRPO (SFT) & 29123 & \underline{40722} & \underline{2.8} & 803 & 14756 & \textbf{22468} & \underline{0.2} & 675 & \underline{4270} & \underline{7751} & \textbf{0.0} & 989 & \textbf{18720} & 39291 & \textbf{0.0} & 1546 & 0.104 & 4.7 \\
    \textbf{LithoGRPO (RLFT)} & \underline{28933} & \textbf{39838} & 3.1 & 444 & 14840 & \underline{22497} & \underline{0.2} & 358 & 4276 & 8376 & \textbf{0.0} & 398 & \underline{19102} & 41359 & \underline{0.2} & 889 & 0.104 & \textbf{4.3} \\
    \quad $\pm$ std dev & $\pm$541 & $\pm$281 & $\pm$0.2 & $\pm$24 & $\pm$643 & $\pm$160 & $\pm$0.1 & $\pm$21 & $\pm$68 & $\pm$122 & $\pm$0.0 & $\pm$31 & $\pm$373 & $\pm$26 & $\pm$0.1 & $\pm$16 & - & - \\
    \bottomrule
  \end{tabular}%
  }
\end{table*}

\begin{figure}[!th]
  \centering
   \includegraphics[width=\columnwidth]{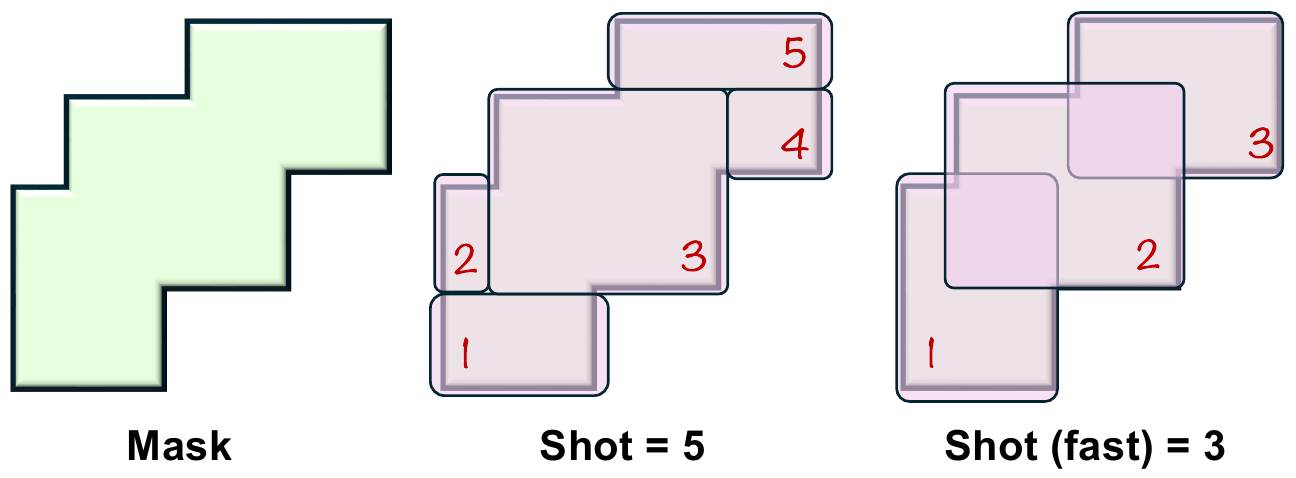}
\caption{\textbf{Shot-counting methods.}
Traditional counting tiles the mask with non-overlapping rectangles; ours uses a minimal overlapping cover, yielding faster computation while largely preserving relative shot-count ordering.}
   \label{fig:shot}
\end{figure}

\textbf{Fast Shot Count.}
The shot count, a critical manufacturability metric, is traditionally obtained by partitioning a mask pattern into the minimum set of \textit{non-overlapping} rectangles.
This minimum-rectangle partitioning problem is NP-hard~\cite{kim2023rectangular}, and existing implementations~\cite{raghuvanshi2009efficient} often take over one minute per mask.
To enable fast evaluation, we compute an \emph{overlapping} shot-count proxy by formulating rectangle selection as a minimum set cover problem and solving it with an ILP (Fig.~\ref{fig:shot}).
Our pipeline has three steps.
(1) Extract a comprehensive set of locally maximal rectangles using a histogram-based search in $O(N^{2})$ for an $N^{2}$-pixel mask, ensuring that every valid shot is contained within at least one candidate.
(2) Redundant rectangles that are fully contained in others are removed via a pruning step in $O(K^{2})$, where $K$ is the number of candidate rectangles.
(3) Remaining rectangles form a compact Set Cover ILP, where a row-wise scan-line optimization generates all covering constraints in $O(NK^{2})$.
This method achieves up to 490$\times$ speedup over traditional non-overlapping partitioning while preserving the relative ordering of the shot count, serving as an efficient proxy.
This approximation is well-suited to GRPO: the group-wise advantage normalization in Eq.~\ref{eq:advantage} cancels any constant offset between $\mathrm{Shot}$(fast) and the traditional shot, so only the within-group ranking of candidates matters for the policy gradient. As Fig.~\ref{fig:shot_correlation} shows, this ranking is strongly preserved.
Moreover, allowing overlapping shots aligns with modern multi-beam mask writing practice~\cite{chua2011optimization, zable2010writing, fujimura2010efficiently, chan2016benchmarking, zhang2024fracturing}.
Details are in Appendix~\ref{sec:details_fast_shot}.

\textbf{Inference.}
During inference, masks are generated by integrating the velocity field using the Euler update in Eq.~\ref{eq:euler}, starting from Gaussian noise $\mathbf{x}_0$. For efficiency, we use a single-step update ($\Delta t\!=\!1$) by default, while multi-step inference is also supported and evaluated in the ablation study.

\section{Experimental Results}

\begin{figure*}[!th]
  \centering
  \includegraphics[width=\textwidth]{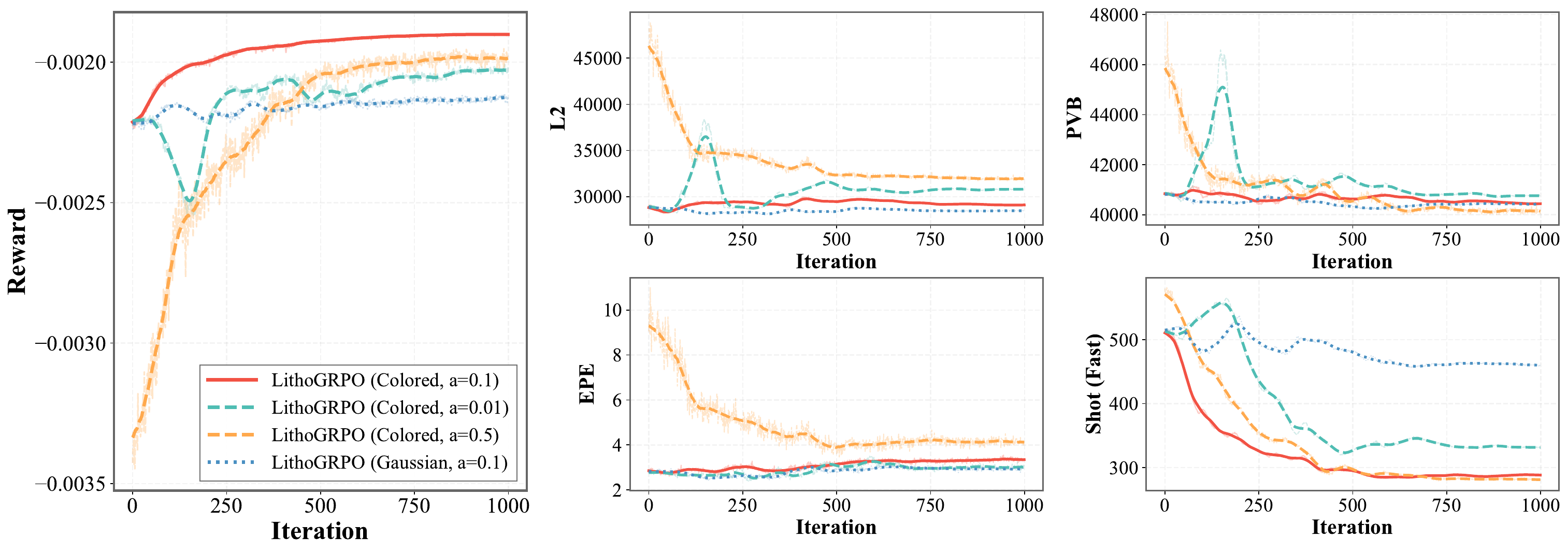}
\caption{\textbf{Noise configurations in SDE.}
We compare noise types (colored vs.\ Gaussian) and noise levels ($a \in \{0.01, 0.1, 0.5\}$) for LithoGRPO.
At $a=0.5$, mask quality degrades, leading to poor initialization; 
at $a=0.01$, exploration slows. 
Gaussian noise yields inferior shot-count results, 
while colored noise with $a=0.1$ achieves the best overall balance.}
  \label{fig:ablation_metalset}
\end{figure*}

\begin{table*}[!ht]
\centering
\caption{\textbf{Effect of inference steps on Metal-layer benchmarks.}
On both MetalSet and its OOD counterpart StdMetal, one-step inference already matches or beats more-step variants on the main metrics while being $2$--$10\times$ faster.
Avg. Rank is computed over 2 datasets $\times$ 4 metrics + Runtime (9 columns), with ties assigned the average of the occupied ranks.
LithoGRPO (RLFT) results are the mean over 4 independent seeds, matching the main table.
}
\label{tab:ablation_timestep}
\small
\begin{tabular}{l|c|rrrr|rrrr|c|c}
\toprule
\multirow{2}{*}{\textbf{Method}} & \multirow{2}{*}{\textbf{Step}} & \multicolumn{4}{c|}{\textbf{MetalSet}} & \multicolumn{4}{c|}{\textbf{StdMetal}} & \multirow{2}{*}{\textbf{\begin{tabular}[c]{@{}c@{}}Runtime\\(s)$\downarrow$\end{tabular}}} & \multirow{2}{*}{\textbf{\begin{tabular}[c]{@{}c@{}}Avg.\\Rank$\downarrow$\end{tabular}}} \\
\cmidrule(lr){3-6} \cmidrule(lr){7-10}
 & & L2 & PVB & EPE & Shot & L2 & PVB & EPE & Shot & & \\
\midrule
ILILT & - & 30353 & 45353 & 3.2 & 433 & 14596 & 24969 & 0.1 & 368 & 0.441 & 7.3 \\
\midrule
\multirow{4}{*}{LithoGRPO (Pretrain)}
 & 1 & 32824 & 42320 & 5.4 & \textbf{487} & 17362 & \textbf{23837} & 0.4 & \textbf{386} & \textbf{0.104} & 9.1 \\
 & 2 & 31540 & \textbf{42167} & 4.2 & 506 & 16404 & 23903 & 0.2 & 402 & 0.203 & 9.1 \\
 & 5 & \textbf{31420} & 42352 & 4.2 & 512 & \textbf{16242} & 23916 & 0.2 & 414 & 0.508 & 9.8 \\
 & 10 & 31790 & 42400 & \textbf{2.8} & 507 & 16330 & 24040 & \textbf{0.2} & 415 & 1.014 & 9.9 \\
\midrule
\multirow{4}{*}{LithoGRPO (SFT)}
 & 1 & 29123 & 40722 & 2.8 & 803 & 14756 & \textbf{22468} & 0.2 & 675 & \textbf{0.104} & 7.4 \\
 & 2 & \textbf{28185} & 40566 & 2.7 & 774 & \textbf{13906} & 22582 & 0.1 & 644 & 0.203 & 5.4 \\
 & 5 & 28520 & \textbf{40456} & \textbf{2.6} & 776 & 14388 & 22486 & 0.1 & 626 & 0.508 & 5.9 \\
 & 10 & 28396 & 40556 & 2.6 & \textbf{758} & 14121 & 22544 & \textbf{0.1} & \textbf{618} & 1.014 & 5.9 \\
\midrule
\multirow{4}{*}{LithoGRPO (RLFT)}
 & 1 & 28933 & \textbf{39838} & 3.1 & \textbf{444} & 14840 & \textbf{22497} & 0.2 & \textbf{358} & \textbf{0.104} & 4.7 \\
 & 2 & \textbf{28386} & 40152 & \textbf{3.0} & 460 & \textbf{13881} & 22743 & \textbf{0.1} & 368 & 0.203 & 3.8 \\
 & 5 & 28630 & 40035 & 3.1 & 483 & 14076 & 22653 & 0.2 & 390 & 0.508 & 5.9 \\
 & 10 & 28658 & 39927 & 3.1 & 491 & 14128 & 22617 & 0.2 & 400 & 1.014 & 6.7 \\
\bottomrule
\end{tabular}%
\end{table*}

\textbf{Experimental Settings.}
We evaluate LithoGRPO and baselines on the LithoBench~\cite{zheng2023lithobench} dataset, which contains metal-layer (MetalSet, StdMetal) and via-layer (ViaSet, StdContact) patterns, each covering a \(2048\times2048~\text{nm}^2\) region. All patterns are processed at \(512\times512\) resolution. 
For metal layers, LithoGRPO is pretrained for 50~epochs on {MetalSet}, followed by 25~epochs of SFT and 1000~RLFT steps, and evaluated on both {MetalSet} and in-distribution {StdMetal}. 
For via layers, 10~epochs of pretraining, 1~epoch of SFT, and 1000~RLFT steps are performed on {ViaSet}; {StdContact}, being out-of-distribution, is fine-tuned from the {ViaSet}-pretrained model for 50~epochs of SFT with 1000~RLFT steps following the LithoBench protocol. 
We evaluate EPE with a 15~nm tolerance, and EPE can saturate at 0 when all edge samples fall within this tolerance.
We use a group size \(G=6\) in GRPO. An 87M-parameter U-Net backbone uses a single denoising step at inference. 
We use PuLP to solve the ILP formulation for fast shot count.
Fast shot count is used for GRPO training, while traditional shot count is reported at inference.
All experiments are conducted on four NVIDIA RTX~3090 GPUs with Intel Xeon Silver 4216 CPUs, with each training stage taking under 8 hours.
Details are in Appendix~\ref{sec:detailed_experimental_settings}.

\textbf{Main Results.}
Table~\ref{tab:main_results_avg_time} compares LithoGRPO with baseline methods. 
Most baseline results are taken from the original papers~\cite{zheng2023lithobench,yang2024ililt}, 
while metrics not reported in the literature (e.g., $\mathrm{Shot}$ for ILILT and optimization-based methods) are reproduced using the same evaluation protocol.
MOSAIC results for the via layer are omitted because its hyperparameters failed to generalize from metal to via patterns, resulting in blank resist images.
The results show that the three training stages of LithoGRPO contribute sequentially: 
Pre-training establishes a basic mask-generation capability,
SFT, which directly optimizes differentiable metrics (\(\mathrm{L2}\) and \(\mathrm{PVB}\)), rapidly reduces them but increases \(\mathrm{Shot}\) complexity,
and RLFT balances all metrics, notably lowering \(\mathrm{Shot}\) while keeping \(\mathrm{L2}\), \(\mathrm{PVB}\), and \(\mathrm{EPE}\) stable.
With one‑step inference, LithoGRPO maintains high efficiency (0.1~s per sample) and achieves the best overall trade‑off among conflicting metrics, attaining the lowest Avg.\ Rank (\textbf{4.3}), ahead of all baselines (best at 5.6), setting a new state of the art.
Additional results are shown in Appendix~\ref{sec:supplementary_results}.

\textbf{Noise Type and Level.}
To enable stochastic exploration in RL, we extend the deterministic ODE sampling (Eq.~\ref{eq:euler}) to an SDE formulation (Eq.~\ref{eq:sde_update_rule_split}). 
The injected noise is controlled by two factors: the noise level $a$ and the noise type (Gaussian white vs.\ low‑frequency colored). 
We analyze their effects on MetalSet in Fig.~\ref{fig:ablation_metalset}.
The default setting is $a=0.1$ with colored noise. 
When decreasing $a$ to $0.01$, the exploration becomes noticeably slower. 
Conversely, increasing $a$ to $0.5$ yields noisier, less stable masks, leading to poorer initial rewards. 
Even under higher noise levels, our method converges to close results, demonstrating robustness.
In contrast, Gaussian noise produces more fragmented masks than low‑frequency colored noise, hindering exploration of lower‑shot mask patterns.

\textbf{Inference Steps.}
Table~\ref{tab:ablation_timestep} compares different rectified flow sampling steps on the Metal benchmarks.
RLFT with two-step sampling attains the best overall rank ($3.8$), narrowly ahead of one-step ($4.7$); the marginal quality gain does not offset the $2\times$ runtime, so one-step sampling is adopted as the default for efficient generation.

\begin{figure}[!ht]
    \centering
    \includegraphics[width=\columnwidth]{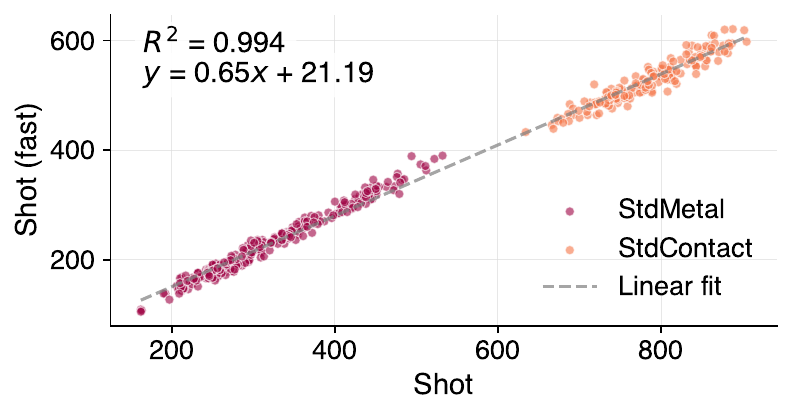}
    \caption{\textbf{Correlation between shot and shot (fast).} 
    Strong linear correlation demonstrates that the fast shot method accurately approximates the traditional shot count.
    }
    \label{fig:shot_correlation}
\end{figure}

\textbf{Fast Shot Count.}
We compare the traditional shot-count implementation in LithoBench with our proposed fast version in correlation and efficiency.
The comparison uses masks generated by LithoGRPO (RLFT). 
As shown in Fig.~\ref{fig:shot_correlation}, the two methods exhibit a strong correlation ($R^2=0.994$), confirming the reliability of our fast estimation scheme.
The ILP formulation is also fully deterministic by construction, yielding perfectly reproducible shot counts.
Table~\ref{tab:runtime_comparison} further reports the runtime, showing a 134$\times$–491$\times$ speed-up over the traditional implementation, greatly improving RL sampling efficiency.
A detailed error analysis between traditional and fast shot count is in Appendix~\ref{sec:fast_shot_estimation}.

\begin{table}[!ht]
\centering
\caption{\textbf{Runtime comparison for shot computation.} 
}
\resizebox{\columnwidth}{!}{%
\begin{tabular}{l|cccc}
\toprule
\textbf{Method} & \textbf{MetalSet} & \textbf{StdMetal} & \textbf{ViaSet} & \textbf{StdContact} \\
\midrule
Shot (s) & 34.01 $\pm$ 8.10 & 64.82 $\pm$ 17.69 & 44.81 $\pm$ 15.88 & 148.35 $\pm$ 14.32 \\
Shot (fast) (s) & \phantom{0}0.25 $\pm$ 0.19 & \phantom{0}0.16 $\pm$ \phantom{0}0.06 & \phantom{0}0.18 $\pm$ \phantom{0}0.12 & \phantom{00}0.30 $\pm$ \phantom{0}0.06 \\
Speedup & $\times$134.6 & $\times$398.2 & $\times$251.1 & $\times$491.3 \\
\bottomrule
\end{tabular}
}
\label{tab:runtime_comparison}
\end{table}

\section{Conclusion}
We present LithoGRPO, the first ILT framework that integrates flow matching with GRPO-based RL fine‑tuning.
It formulates mask generation as a rectified flow, enabling the joint optimization of differentiable and non‑differentiable metrics via reward feedback.
A fast shot‑count algorithm further accelerates mask evaluation by over 130×, ensuring efficient exploration.
Experiments demonstrate high accuracy, robustness, manufacturability, and efficiency.

\textbf{Limitations.}
Our multi-stage training adds computation but yields the best overall trade-off, and jointly optimizing conflicting metrics remains inherently challenging (Table~\ref{tab:main_results_avg_time}). Evaluation is limited to public academic benchmarks (LithoBench); industrial-scale validation and extreme layout densities are left for future work.

\section*{Impact Statement}

This paper presents work whose goal is to advance the field of Machine Learning.
Our experiments rely on open-source academic benchmarks at mature technology nodes, far from cutting-edge industrial processes.
We do not foresee additional societal consequences requiring specific discussion.

\section*{Acknowledgment}
This paper is partially supported by the National Key R\&D Program of China No.2022ZD0161000 and the General Research Fund of Hong Kong No.17200622 and 17209324.

\bibliography{example_paper}
\bibliographystyle{icml2026}

\newpage
\appendix
\onecolumn

\section{Related Work}

\subsection{Optical Proximity Correction}
\begin{figure}[!ht]
  \centering
  \includegraphics[width=0.7\columnwidth]{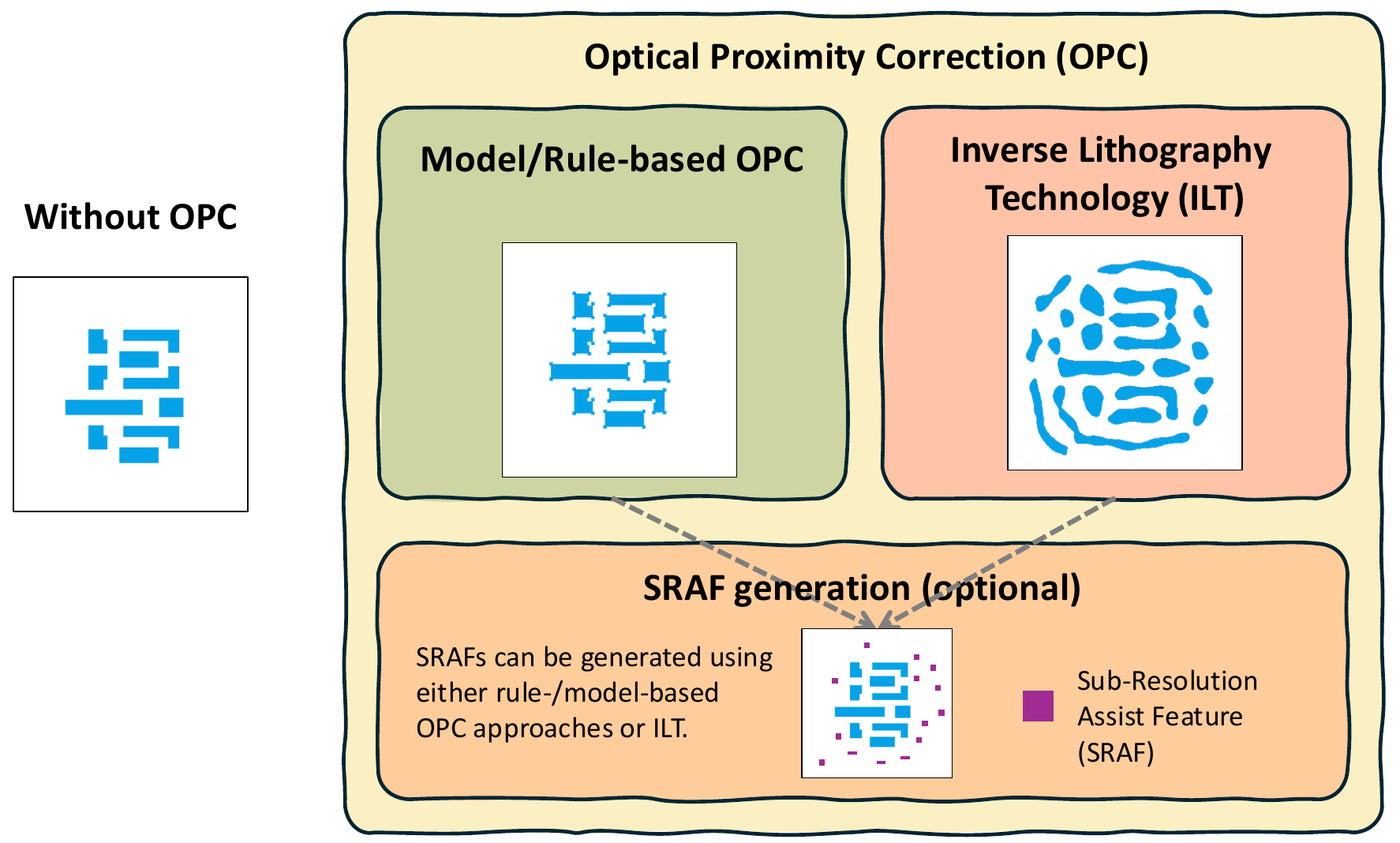}
  \caption{\textbf{Relationship between OPC, ILT, and optional SRAF insertion/generation.}}
   \label{fig:taxnomony}
\end{figure}

Optical proximity correction (OPC)~\cite{levinson2005principles, moreau2012semiconductor, mack2008fundamental, jin2025recent} is the mainstream mask correction technique in optical lithography to compensate for imaging distortions and process variations. 
Rule-based OPC~\cite{otto1994automated, dong2017analog, weng2024enhanced, liang2024camo} applies pre-defined geometric recipes (e.g., bias tables and pattern-dependent corrections) to modify mask features, offering high throughput but limited fidelity under complex imaging conditions. 
Model-based OPC~\cite{wu2024method, zhu2025machine, sun2023efficientiseda, geng2019sraf} uses calibrated lithography and resist models to predict wafer contours and iteratively updates mask edges to minimize objective functions such as EPE and CD errors, achieving higher accuracy at the cost of heavier computation. 
Sub-resolution assist features (SRAFs)~\cite{alawieh2019gan, xu2016machine, liu2022sub} are commonly inserted as a resolution enhancement technique to improve image contrast and process window; in practice, SRAF insertion can be performed as a standalone step or co-optimized within the OPC flow. 
In contrast to edge-centric OPC, inverse lithography technology (ILT) formulates mask synthesis as a global optimization problem and can naturally generate complex mask geometries, including SRAF-like patterns~\cite{yu2023ctm}, to better satisfy lithographic objectives.
Beyond mask optimization, learning-based methods have also been explored to model and generate photolithography overlay maps for process monitoring and correction~\cite{yang2025photolithography}.
We give an overview of the relationship between traditional OPC, ILT, and SRAF in Fig.~\ref{fig:taxnomony}.

\subsection{Inverse Lithography Technology}

ILT approaches can be broadly categorized into two paradigms. Optimization-based methods~\cite{gao2014mosaic, yu2022gpu, sun2023efficient, sun2025adaptive, jia2010machine, luo2024enabling} iteratively refine mask patterns through gradient-based optimization of differentiable metrics such as L2 distance and process variation band (PVB). Representative works include MOSAIC~\cite{gao2014mosaic}, which incorporates process window awareness, and GPU-accelerated level-set methods~\cite{yu2022gpu} that improve computational efficiency. Multi-level simulation approaches~\cite{sun2023efficient, sun2025adaptive} further enhance optimization speed while maintaining accuracy. However, these methods are limited to differentiable objectives and require multiple costly lithography simulations.

Learning-based approaches~\cite{yang2018gan, chen2020damo, jiang2020neural, jiang2021neural, yang2022generic, yang2022large, zhu2023l2o, wu2025tokman, wu2025lvmmo, liang2024rlopc} leverage neural networks to learn direct mappings from target patterns to masks, significantly accelerating mask generation. GAN-OPC~\cite{yang2018gan} pioneered the use of generative adversarial networks for mask optimization, while Neural-ILT~\cite{jiang2020neural, jiang2021neural} and DAMO~\cite{chen2020damo} demonstrated full-chip scalability. More recent works incorporate domain-specific architectures~\cite{yang2022generic}, learning-to-optimize frameworks~\cite{zhu2023l2o}, tokenization-based approaches with manufacturing constraints~\cite{wu2025tokman}, and large vision models for full-chip optimization~\cite{wu2025lvmmo}. 
Diffusion models have also been proposed for ILT tasks, such as HiFiMo~\cite{haoyu2025hifimo} and BBDM~\cite{haoxiang2025bbdm}.
However, these methods only optimize the L2 loss and PVB metric rather than all evaluation objectives. Moreover, their efficiency is still constrained by the number of inference steps required in diffusion.
The LithoBench benchmark~\cite{zheng2023lithobench} provides standardized evaluation protocols for comparing ILT methods. Recent hybrid approaches~\cite{yang2024ililt} attempt to combine optimization and learning, though they remain constrained by computational overhead and inability to optimize non-differentiable metrics such as shot count and edge placement error (EPE).

Reinforcement learning has been explored for mask optimization. RL-OPC~\cite{liang2024rlopc} pioneered the use of deep reinforcement learning in optical proximity correction, demonstrating the capability of RL to directly optimize non-differentiable objectives without requiring differentiable proxies. However, RL-OPC operates within the traditional edge-based OPC framework, performing local adjustments to individual mask features rather than holistic mask generation. This edge-centric approach limits its optimization scope to local neighborhoods and makes it challenging to simultaneously balance conflicting metrics across the entire layout. Furthermore, traditional shot count computation remains a bottleneck for RL-based optimization, as the NP-hard non-overlapping rectangle partitioning problem requires prohibitive computation time. Our work addresses these limitations by formulating generative ILT within a flow-matching framework and introducing an ultra-fast shot count estimation method that enables efficient GRPO-based reinforcement learning fine-tuning.
Concurrent work~\cite{yang2026pushing} also applies GRPO to ILT, but uses a WGAN generator refined by a post-hoc numerical solver at inference, whereas our method generates masks directly with one-step rectified flow.

Mask manufacturability considerations, particularly shot count optimization, have been extensively studied in the EDA community~\cite{fujimura2010efficiently, zable2010writing, chua2011optimization, chan2016benchmarking, zhang2024fracturing}. These works establish the importance of efficient mask fracturing and shot placement for multi-beam mask writers, motivating the development of fast shot count estimation methods for computational lithography.

\subsection{GRPO for Flow-Matching Models}
Diffusion and flow-matching models~\cite{ho2020denoising, song2020denoising, rombach2022high, lipman2023flow} decompose visual generation into iterative denoising processes, significantly advancing visual synthesis and achieving state-of-the-art performance in image and video generation. Inspired by the success of reinforcement learning (RL) in large language models (LLMs), optimization techniques such as PPO~\cite{black2023training} and DPO~\cite{wallace2024diffusion} have been adapted to diffusion models, facilitating preference alignment and enhancing task-specific outcomes. In a similar vein, Flow-GRPO~\cite{liu2025flow} and DanceGRPO~\cite{xue2025dancegrpo} incorporate GRPO-style policy optimization into flow-matching frameworks by reformulating deterministic ODE sampling as stochastic SDE processes, thereby introducing exploratory noise for group-based policy improvement. More recently, MixGRPO~\cite{li2025mixgrpo} introduced a hybrid ODE–SDE sampling strategy that enhances training efficiency without compromising generative quality. Concurrently, Flow-CPS~\cite{wang2025coefficients} identified a critical limitation in the SDE sampling employed by Flow-GRPO and DanceGRPO—inconsistent noise coefficients across timesteps—which results in residual noise accumulation and imprecise reward estimation. To mitigate this, Flow-CPS proposes a noise-consistent SDE sampling method that improves reward accuracy and accelerates GRPO convergence.
In parallel, TempFlowGRPO~\cite{he2025tempflow} and G$^2$RPO~\cite{zhou2025text} tackle the issues of reward sparsity and inaccuracy arising from assigning a single global reward to multi-step SDE trajectories.
Along the line of addressing sparse/ambiguous supervision over multi-step trajectories, E-GRPO~\cite{zhang2026grpo} identifies that only high-entropy steps contribute to effective exploration, and proposes entropy-aware step consolidation with a multi-step group-normalized advantage to improve learning efficiency.
BranchGRPO~\cite{li2025branchgrpo} reorganizes the rollout process into a branching tree structure, where shared prefixes reduce computational overhead and pruning eliminates low-reward paths and redundant depths. 
GRPO-Guard~\cite{wang2025grpo} introduces a regulated clipping mechanism that stabilizes the optimization process and substantially alleviates implicit over-optimization, achieving this without relying on strong KL regularization.
Beyond improvements in sampling and credit assignment, recent GRPO-based visual alignment work has begun to explicitly address diversity degradation and reward hacking.
DiverseGRPO~\cite{liu2025diversegrpo} mitigates mode collapse by adding a distribution-level creativity bonus via semantic grouping and by applying structure-aware regularization that emphasizes early denoising to preserve diversity.
GARDO~\cite{he2025gardo} tackles reward hacking from imperfect proxy rewards through gated, uncertainty-aware regularization with an adaptively updated reference, while further encouraging mode coverage with diversity-aware reward shaping.
Multi-GRPO~\cite{lyu2025multi} improves sparse-reward credit assignment and multi-objective stability by using tree-based temporal grouping for advantage estimation and reward-wise grouping for decoupled advantage computation before aggregation.

\section{Additional Details on Fast Shot Count }
\label{sec:details_fast_shot}

We provide a detailed breakdown of the algorithm used for ultra-fast shot count estimation. The method transforms the computationally expensive non-overlapping rectangle partitioning problem into an efficient Set Cover problem solved via Integer Linear Programming (ILP). The overall process is presented in Algorithm~\ref{alg:fast_shot_count}.

\subsection{Overview}

Given a binary mask $M$ of size $N \times N$ ($N=512$ in our LithoGRPO), we aim to find the minimum number of rectangular shots (which may overlap) that cover all pixels with value 1. The algorithm consists of three main steps:

\begin{enumerate}
    \item \textbf{Maximal Rectangle Generation}: Extract all maximal rectangles from the binary mask in $O(N^2)$ time.
    \item \textbf{Redundancy Pruning}: Remove rectangles that are fully contained within others, reducing the problem size.
    \item \textbf{ILP Formulation}: Formulate a Set Cover ILP with scan-line optimization.
    \item \textbf{ILP Solve}: Solve the ILP task with solver.
\end{enumerate}

\subsection{Step 1: Maximal-Rectangle Generation}
\label{sec:maxrect}

\begin{figure}[!ht]
    \centering
    \includegraphics[width=\linewidth]{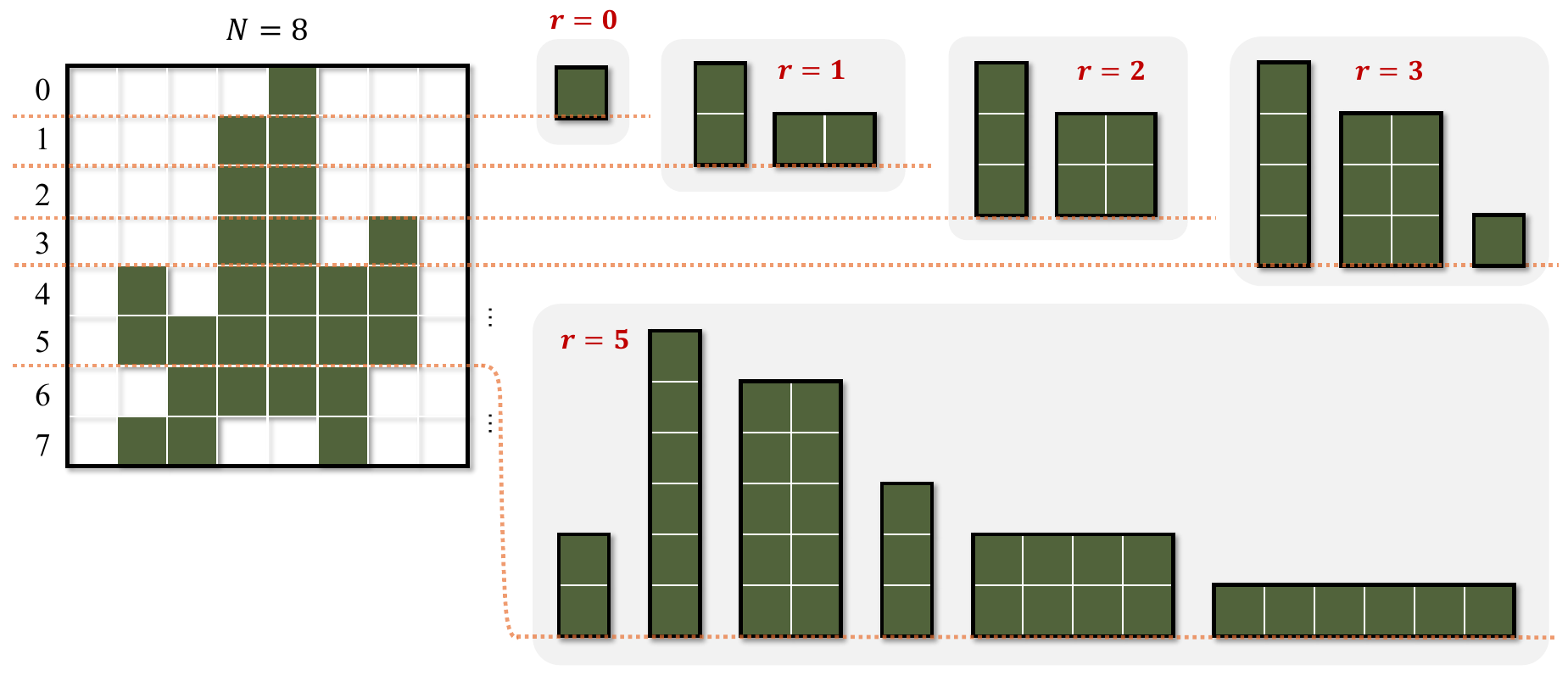}
   \caption{\textbf{Visualization of maximal-rectangle generation.}
Left: an $8\times 8$ binary mask, where colored (filled) pixels indicate $1$'s.
Right: maximal rectangles enumerated for selected rows $r\in\{0,1,2,3,5\}$, where each rectangle's bottom edge lies on row $r$.}
    \label{fig:maximal-rectangle}
\end{figure}

A \emph{maximal rectangle} is an axis-aligned rectangle that consists
exclusively of foreground pixels ($1$’s) and cannot be enlarged in any
direction without hitting background ($0$’s).
We enumerate all such rectangles with a histogram sweep that scans the
mask once:

\begin{enumerate}
    \item For every row $r\in\{0,\dots,N-1\}$ maintain a height array
          $h[c]$ that counts the number of consecutive $1$’s directly
          above (and including) pixel $(r,c)$.
    \item Treat $h[\cdot]$ as a histogram and, in $O(N)$ time,
          enumerate \emph{all} maximal rectangles whose bottom edge lies
          on row $r$ using the classic monotonic-stack algorithm for the
          ``largest rectangle in a histogram'' problem. For example, in an $8\times 8$ binary mask, the maximal rectangles for $r \in \{0,1,2,3,5\}$ are shown in Fig.~\ref{fig:maximal-rectangle}.
    \item Append every rectangle found in the second step to the candidate set
          $\mathcal{R}_{\text{all}}$; let
          $K_{\text{all}}=|\mathcal{R}_{\text{all}}|$.
\end{enumerate}

Because each of the $N$ rows is processed once, and the per-row work is
linear in the image width, the total running time of this stage is $O(N^2)$.

The set $\mathcal{R}_{\text{all}}$ is \emph{complete}: any feasible shot
rectangle must be contained in at least one element of
$\mathcal{R}_{\text{all}}$.  Subsequent stages only remove redundancies.

\subsection{Step 2: Redundancy Pruning}
The raw candidate set $\mathcal{R}_{\mathrm{all}}$ still contains many
rectangles that do not need to appear in the final covering.
A rectangle $A$ is redundant if it is fully contained in another rectangle
$B$ because any solution that uses $A$ can always replace it by $B$ without
increasing the objective.

We employ a Numba-accelerated~\cite{lam2015numba} greedy strategy:

\begin{enumerate}
    \item Sort all rectangles by area in descending order.
    \item For every rectangle, test whether it is
          contained in any rectangle that has already been kept.
          
    \item Append the rectangle to the pruned set
          $\mathcal{R}_{\mathrm{pruned}}$ iff it is not contained.
\end{enumerate}

After this step, we obtain a pruned set of rectangles, denoted by $\mathcal{R}_{\mathrm{pruned}}$, and define $K = \lvert \mathcal{R}_{\mathrm{pruned}} \rvert$.
The procedure needs
$\mathcal O\!\bigl(K_{\text{all}}^{\;2}\bigr)$
comparisons in the worst case, yet it is fast in practice because
$K_{\text{all}}\!\ll N^{2}$ for real masks.
For extremely large instances, one can switch to the
$\mathcal O(K_{\text{all}}\log^{2}K_{\text{all}})$
variant based on a 2-D Binary Indexed Tree, but we found its overhead
unnecessary for our datasets.

\subsection{Step 3: ILP Formulation with Scan-Line Constraint Generation}

Let $x_i\in\{0,1\}$ indicate whether pruned rectangle
$R_i\in\mathcal{R}_{\mathrm{pruned}}$ is selected. 
Let $M:\{0,\dots,H-1\}\times\{0,\dots,W-1\}\to\{0,1\}$ be the binary target mask, where $M(p)=1$ indicates that pixel $p$ belongs to the target shape (i.e., is to be covered).

The covering ILP is

\begin{align}
    \text{minimise}\quad & \sum_{i=1}^{K} x_i\\
    \text{subject to}\quad &
    \sum_{i:\;p\in R_i} x_i \;\ge 1,
    \qquad \forall\,p\text{ with }M(p)=1 ,
\end{align}

where $K=|\mathcal{R}_{\mathrm{pruned}}|$.
Creating one constraint per target pixel would yield up to $\Theta(N^2)$ constraints, which is computationally prohibitive. We therefore apply a scan-line reduction:

\begin{enumerate}
    \item For each row $r$, compute the run-length encoding (RLE) of $M(r,\cdot)$, i.e., the set of maximal contiguous foreground segments $(s,e)$ such that $M(r,c)=1$ for all $c\in[s,e]$. For example, if $M(r,\cdot)=0011110011100$, then the RLE segments are $(s,e)=(2,5)$ and $(8,10)$ (0-indexed).
    \item For each row $r$ collect the rectangles that intersect it,
          $\mathcal{R}_r=\{R_i\mid R_i\cap\text{row }r\neq\varnothing\}$.
   \item Collect the \emph{critical} $x$-coordinates in row $r$ at which coverage can change:
the left/right $x$-boundaries of all rectangles in $\mathcal{R}_r$,
together with the endpoints $s,e$ of the foreground RLE segments in row $r$.
For example, with RLE segments $(2,5)$ and $(8,10)$ and rectangles spanning
$[1,6)$, $[4,9)$, and $[8,12)$ on row $r$, we collect
$\{1,6,4,9,8,12\}\cup\{2,5,8,10\}$.

\item Sort and deduplicate the critical coordinates. Consecutive pairs define
intervals on which the set of covering rectangles is constant.
For the example above, sorting yields $\{1,2,4,5,6,8,9,10,12\}$ and thus the
intervals $[1,2),[2,4),[4,5),\ldots,[10,12)$.
\item For every interval that overlaps foreground pixels, insert one covering
constraint using any representative column $c$ inside that interval:
$
\sum_{i:\;c\in R_i} x_i \ge 1.
$
For example, if an interval is $[4,5)$ and on row $r$ it is covered by exactly
$R_1$ and $R_2$, we may choose $c=4$ and add $x_1+x_2\ge 1$.
\end{enumerate}

This replaces per-pixel covering constraints with per-interval constraints
without changing the objective or the feasible set: within a fixed row $r$,
the set $\{i \mid c\in R_i\}$ is constant for all columns $c$ inside the same
interval between consecutive critical $x$-coordinates, so a single constraint
at any representative column $c$ enforces coverage for every foreground pixel
in that interval.

For a fixed row $r$, we collect $\Theta(k_r+t_r)$ critical $x$-coordinates
(rectangle boundaries and RLE endpoints) and sort them, which costs
$\mathcal{O}\bigl((k_r+t_r)\log(k_r+t_r)\bigr)$ time. The sorted coordinates
induce $\Theta(k_r+t_r)$ intervals; to generate one covering constraint per
foreground-overlapping interval we test membership against the $k_r$ rectangles
intersecting the row, costing $\mathcal{O}(k_r)$ per interval and thus
$\mathcal{O}\bigl(k_r(k_r+t_r)\bigr)$ per row. Summing the per-row costs over
$r=1,\dots,N$ yields
\[
\sum_{r=1}^{N}\mathcal{O}\!\Bigl((k_r+t_r)\log(k_r+t_r)+k_r(k_r+t_r)\Bigr)
=\sum_{r=1}^{N}\mathcal{O}\!\bigl(k_r^{2}+k_r\log(k_r+t_r)\bigr),
\]
where the last step groups lower-order terms for readability, and  $k_r$ is the number of pruned rectangles intersecting row $r$ and $t_r$
is the number of foreground RLE runs in that row (for the example above,
$k_r=3$ and $t_r=2$).
In the worst case
$k_r=\Theta(K)$ for all rows, yielding the bound
$\mathcal{O}(NK^{2}+NK\log K)$. In practice, masks exhibit spatial
locality so $k_r\ll K$ for most rows; consequently the quadratic term
$\sum_r k_r^{2}$ is much smaller than the worst-case $NK^{2}$, and the runtime
is often dominated by sorting critical points, i.e.,
$\sum_r \mathcal{O}\!\bigl(k_r\log(k_r+t_r)\bigr)$.

\begin{figure}[t]
\centering
\begin{minipage}{0.6\linewidth}
\begin{algorithm}[H]
\caption{Fast Shot Count Estimation}
\label{alg:fast_shot_count}
\begin{algorithmic}[1]
\REQUIRE Binary mask $M\in\{0,1\}^{N\times N}$
\ENSURE Minimum overlapping shot count $S$
\STATE \textbf{FastShotCount}($M$):
\STATE {\bf Step 1: Maximal-Rectangle Generation}
\STATE $\mathcal{R}_{\mathrm{all}} \gets \textsc{MaximalRectangles}(M)$
\STATE {\bf Step 2: Redundancy Pruning}
\STATE $\mathcal{R}_{\mathrm{pruned}} \gets \textsc{PruneContained}(\mathcal{R}_{\mathrm{all}})$
\STATE $K \gets |\mathcal{R}_{\mathrm{pruned}}|$
\STATE {\bf Step 3: ILP Formulation with Scan-Line Constraints}
\STATE create binary vars $x_1,\dots,x_K$, objective $\min \sum_i x_i$
\STATE $\text{runs} \gets \textsc{RunLengthEncode}(M)$
\FOR{$r \gets 0$ \textbf{to} $N-1$}
  \STATE $\mathcal{R}_r \gets \{ i \in \{1,\dots,K\} \mid R_i \text{ intersects row } r \}$
  \STATE $C \gets \textsc{CriticalPoints}(\mathcal{R}_r,\text{runs}[r])$
  \STATE sort and deduplicate $C$ ascending
  \IF{$|C| < 2$}
    \STATE \textbf{continue}
  \ENDIF
  \FOR{$j \gets 1$ \textbf{to} $|C|-1$}
    \STATE $x_L \gets C_j,\quad x_R \gets C_{j+1}$
    \IF{$\textsc{OverlapsForeground}(\text{runs}[r], [x_L,x_R))$}
      \STATE $c \gets x_L$
      \STATE $\mathcal{I} \gets \{ i \in \mathcal{R}_r \mid c \in R_i \}$
      \IF{$\mathcal{I}=\varnothing$}
        \STATE \textbf{return} infeasible
      \ENDIF
      \STATE add constraint $\sum_{i\in\mathcal{I}} x_i \ge 1$
    \ENDIF
  \ENDFOR
\ENDFOR
\STATE {\bf Step 4: ILP Solve}
\STATE $\text{solution} \gets \textsc{SolveILP}()$
\STATE $S \gets \sum_{i=1}^{K} \text{solution}(x_i)$
\STATE \textbf{return} $S$
\end{algorithmic}
\end{algorithm}
\end{minipage}
\end{figure}

\subsection{Step 4: ILP Solve}
The final ILP has $K$ binary variables and $C$ covering constraints.
Let $\bar{k}$ denote the average number of rectangles covering a foreground
interval (equivalently, the average number of nonzeros per covering constraint).
Due to limited overlap, $\bar{k}$ is small in practice, so the constraint
matrix is sparse. 
Consequently, the resulting ILP instances are typically
solved quickly by a modern solver. We formulate the ILP using the PuLP 
package and solve it with the CBC solver.

\subsection{Computational Complexity}
\label{sec:complexity}

Let $N$ denote the image size ($H=W=N$). Let $K_{\text{all}}$ be the number of
locally maximal rectangles produced in Step~1, and let $K\le K_{\text{all}}$
be the number remaining after redundancy pruning. For each row $r$, let $k_r$
be the number of pruned rectangles intersecting row $r$, and let $t_r$ be the
number of foreground RLE runs in that row. Note that $\sum_{r=1}^N k_r \le KN$
(each rectangle intersects at most $N$ rows) and $\sum_{r=1}^N t_r\le N\lceil N/2\rceil
=\mathcal O(N^2)$.

\begin{enumerate}
\item \textbf{Maximal-rectangle enumeration.}
We compute the maximal rectangles in $\mathcal O(N^2)$ time:
\[
T_1=\mathcal O(N^2).
\]

\item \textbf{Redundancy pruning.}
Sorting by area and greedy pairwise containment tests yield
\[
T_2=\mathcal O(K_{\text{all}}^{\,2}).
\]

\item \textbf{Scan-line constraint generation.}
For each row $r$, we collect $\Theta(k_r+t_r)$ critical $x$-coordinates and
sort them in $\mathcal O((k_r+t_r)\log(k_r+t_r))$ time. These points define
$\Theta(k_r+t_r)$ intervals. For every interval overlapping foreground pixels,
we add one covering constraint; in the simplest implementation, building each
constraint scans the $k_r$ rectangles intersecting the row, giving
$\mathcal O(k_r(k_r+t_r))$ time per row. Summing over rows,
\[
T_3=\sum_{r=1}^{N}\mathcal O\!\Bigl((k_r+t_r)\log(k_r+t_r)+k_r(k_r+t_r)\Bigr).
\]
Using $k_r\le K$ and $t_r\le \lceil N/2\rceil$ for all $r$ gives the coarse
worst-case bound
\[
T_3=\mathcal O\!\bigl(N(K+N)\log(K+N)+NK^2+N^2K\bigr).
\]
In particular, when $K\ge N$ this simplifies to $T_3=\mathcal O(NK^2)$.
\end{enumerate}

Therefore,
\[
T=\mathcal O\!\bigl(N^2+K_{\text{all}}^{\,2}+NK^2\bigr).
\]

In typical masks, spatial locality makes $k_r$ small for most rows, so the observed runtime is closer to
$\sum_r \mathcal O((k_r+t_r)\log(k_r+t_r))$ plus a modest scanning cost.
By contrast, a naive per-pixel approach has $\Theta(|\Omega|)$ constraints and
can take $\Theta(|\Omega|K)$ time just to construct them (worst-case
$\Theta(N^2K)$), whereas our scan-line method constructs far fewer constraints.

\subsection{Traditional Shot Count Analysis}
According to the practical implementation in LithoBench~\cite{zheng2023lithobench}, the shot count is computed using the \texttt{adaptive-boxes} (AdaBox) package, which greedily decomposes the foreground pixels into a set of axis-aligned rectangles.

Given a binary mask, AdaBox first represents all foreground pixels as a point set and then iteratively generates rectangles until all points are covered. In each iteration, it samples several seed points from the currently uncovered set, grows a candidate rectangle around each seed by expanding along the seed's row/column, selects the candidate with the largest area, and finally marks all points inside the chosen rectangle as covered. The number of selected rectangles is returned as the shot count.

This procedure is slow in our setting, mainly due to repeated point-set scans. Evaluating a single candidate rectangle involves filtering the uncovered point array to collect points on the seed's row/column (followed by sorting and boundary extraction), which costs $\Theta(M)$ time up to logarithmic factors, where $M$ is the number of foreground pixels. After a rectangle is selected, AdaBox updates the global assignment by testing all points against the rectangle bounds, incurring another $\Theta(M)$ full pass. Let $S$ denote the number of random seeds evaluated per iteration; LithoBench uses $S=4$. Let $R$ be the number of rectangles produced by the greedy procedure, i.e., the resulting (reported) shot count. A coarse runtime estimate is therefore
$$
T_{\text{AdaBox}} \approx \Theta\!\bigl(R\,(S\,M + M)\bigr)=\Theta(RSM),
$$
again up to sorting-related logarithmic factors. 
Since $M \le N^2$, this implies a worst-case bound $T_{\text{AdaBox}}=\mathcal{O}(RSN^2)$ when the foreground occupies a constant fraction of an $N\times N$ mask. 
Empirically, our masks are often fragmented and AdaBox typically outputs $R=\mathcal{O}(10^2\text{--}10^3)$ rectangles, so the repeated $\Theta(M)$ scans dominate and can lead to minute-level runtimes on $512\times512$ instances even with the small constant $S=4$.

In contrast, our fast shot count formulates the problem as an Integer Linear Program (ILP) over a carefully chosen set of candidate rectangles. A naive approach would enumerate all possible axis-aligned rectangles contained in the foreground region, yielding $\mathcal{O}(N^4)$ candidates in the worst case, which is far too many for practical ILP solving. However, we observe that when rectangles are allowed to overlap, any feasible cover using a non-maximal rectangle can be replaced by one using a maximal rectangle that contains it, without increasing the objective. This dominance property allows us to restrict the candidate set to \emph{maximal rectangles} only, i.e., rectangles that cannot be extended in any direction while remaining inside the foreground. Using a standard histogram-based technique, we enumerate all maximal rectangles in $\mathcal{O}(N^2)$ time: for each row, we maintain the height of consecutive foreground pixels above it, then apply a 
monotonic stack to extract all locally maximal rectangles ending at 
that row. This typically produces $K = \mathcal{O}(10^3\text{--}10^4)$ 
candidates.

To further reduce the ILP size, we employ scan-line aggregation: instead of generating one covering constraint per foreground pixel, we observe that pixels sharing the same row and covered by identical sets of candidate rectangles can be merged into a single constraint. This reduces the number of constraints from $\mathcal{O}(M)$ to $\mathcal{O}(K)$ in practice. The resulting ILP has $K$ binary variables and only a few thousand constraints, which modern solvers such as CBC handle in sub-second time.

Therefore, the speedup primarily comes from (1) restricting candidates to maximal rectangles via the overlap-dominance argument, (2) avoiding $R$ rounds of point-set filtering and global re-scans, and (3) replacing per-pixel constraints with aggregated scan-line constraints, yielding a compact optimization problem solvable in a single pass.

Additionally, due to its greedy nature with random seed sampling, AdaBox produces non-deterministic results: running the same mask multiple times may yield different shot counts. In contrast, our ILP-based formulation is fully deterministic: given the same mask, it always returns the same optimal shot count, which is desirable for reliable evaluation and reproducibility.

\section{Log-Probability Computation}
\label{sec:log_prob_details}

For GRPO optimization, we compute the log-probability of a sampled trajectory as the sum 
of per-step transition log-probabilities, following Flow-GRPO~\cite{liu2025flow}:
\begin{equation}
\log \pi_\theta(\mathbf{x}) = \sum_{t} \log p(\mathbf{x}_{t+\Delta t} | \mathbf{x}_t).
\end{equation}
Each transition follows a Gaussian distribution:
\begin{equation}
\log p(\mathbf{x}_{t+\Delta t} | \mathbf{x}_t) \propto 
-\frac{\|\mathbf{x}_{t+\Delta t} - \boldsymbol{\mu}_t\|^2}{2\sigma_t^2 \Delta t},
\end{equation}
where $\boldsymbol{\mu}_t = \mathbf{x}_t + \big[ \mathbf{v}_{\theta}(\mathbf{x}_t, t) 
+ \frac{\sigma_t^2}{2t}( \mathbf{x}_t + (1-t)\mathbf{v}_{\theta}(\mathbf{x}_t, t) ) \big]\Delta t$ 
is the deterministic component in Eq.~\ref{eq:sde_update_rule_split}, 
and we assume pixel-wise independence for computational efficiency.

Although the colored noise used in our SDE sampler introduces spatial correlations between pixels, 
we approximate the log-probability using an independent Gaussian. 
This approximation is valid for GRPO because only the probability ratio 
$\pi_\theta / \pi_{\text{ref}}$ is used, and both policies share the same noise generation process, 
causing the approximation error to largely cancel out.

\section{Model Architecture}
\label{sec:model_architecture}

\noindent\textbf{Overall Structure.}
Our model adopts a symmetric encoder–decoder architecture summarized in Table~\ref{tab:miniunet_architecture}.
Each downsampling stage halves the spatial resolution and doubles the channel dimension,
while the decoder mirrors this hierarchy with upsampling and skip connections.
All convolutions use kernel size $3\times3$, stride~1, and padding~1.
Group normalization (8 groups) and SiLU activation follow each convolution.
The network takes a two‑channel input (layout and mask) and outputs a single‑channel prediction.
The base channel width is~64, and it doubles after each down-sampling stage, 
reaching 1024 channels at the bottleneck.

\begin{table}[!ht]
\centering
\caption{\textbf{Model architecture.}
The network takes a \textbf{two‑channel input}, where the first channel is the noisy sample $x_t$ 
and the second is the conditional target (e.g., mask or layout).
Each DownLayer and UpLayer is a residual block with two $3\times3$ convolutions,
GroupNorm (8 groups), and SiLU activation.
Skip connections concatenate the encoder and decoder features of matching resolutions.}
\begin{tabular}{lccc}
\toprule
\textbf{Stage} & \textbf{Layer(s)} & \textbf{Channels} & \textbf{Output Size} \\
\midrule
Input & Conv2d, $3\times3$, padding=1 & 2→64 & $256\times256$ \\
Down1 & 2×DownLayer; MaxPool($2\times2$) & 64→128 & $128\times128$ \\
Down2 & 2×DownLayer; MaxPool($2\times2$) & 128→256 & $64\times64$ \\
Down3 & 2×DownLayer; MaxPool($2\times2$) & 256→512 & $32\times32$ \\
Down4 & 2×DownLayer; MaxPool($2\times2$) & 512→1024 & $16\times16$ \\
Middle & MiddleLayer & 1024→1024 & $16\times16$ \\
Up1 & Upsample($\times2$); 2×UpLayer & 2048→512 & $32\times32$ \\
Up2 & Upsample($\times2$); 2×UpLayer & 1024→256 & $64\times64$ \\
Up3 & Upsample($\times2$); 2×UpLayer & 512→128 & $128\times128$ \\
Up4 & Upsample($\times2$); 2×UpLayer & 256→64 & $256\times256$ \\
Output & Conv2d, $1\times1$ & 64→1 & $256\times256$ \\
\bottomrule
\end{tabular}
\label{tab:miniunet_architecture}
\end{table}

\paragraph{Layer Modules.}
Each DownLayer consists of two $3\times3$ convolutions (stride 1, padding 1),
each followed by GroupNorm (8 groups) and SiLU activation.
A residual shortcut adds the input to the output; when channel dimensions differ,
a $1\times1$ convolution aligns them.
A linear projection of the time embedding 
($t_{\mathrm{emb}}\!\in\!\mathbb{R}^{B\times d}$, 
where each of the $B$ samples has a $d$‑dimensional time embedding)
is added to the feature maps before the first convolution.
DownLayers are applied before each $2\times2$ max‑pooling operation in the encoder.

The UpLayer mirrors the DownLayer structure.
Each block first upsamples features by a factor of 2 using nearest‑neighbor interpolation,
then applies two $3\times3$ convolutions with GroupNorm and SiLU activations, plus the same residual connection and time‑embedding injection as in the encoder.

The MiddleLayer has the same residual structure as the DownLayer
but operates at the bottleneck resolution without pooling or upsampling.
It refines the deepest latent features before entering the decoder.

\paragraph{Normalization and Activation.}
All normalization layers use GroupNorm with eight groups,
and all nonlinearities employ the SiLU activation function.

\paragraph{Time Embeddings.}
Time embeddings are encoded using sinusoidal positional encodings
and projected through a learnable linear layer to match the feature‑map dimensionality. The resulting embedding is added element‑wise to feature maps at each block, allowing the network to incorporate temporal information during training.

\section{Detailed Experimental Settings}
\label{sec:detailed_experimental_settings}

We report the training and inference settings in our experiments in Table~\ref{tab:exp_settings}. For reinforcement learning fine‑tuning (RLFT), the model is optimized for 1000 iterations rather than epochs. Each iteration uses 10 images as input, and for every image, $G=6$ exploratory samples are generated, forming a mini‑batch of 60 samples for one gradient update.

For learning rate scheduling, different strategies are employed for each stage: (1) during pretraining, we use a StepLR scheduler that decays the learning rate by a factor of 0.1 halfway through training; (2) during RLFT, we apply a warm‑up with cosine decay scheduler, where the first 5\% of iterations are used for linear warm‑up and the learning rate gradually decreases to 5\% of the maximum; (3) no learning rate scheduling is applied during SFT.

\begin{table}[!ht]
\centering
\caption{\textbf{Training and inference settings.}
Each column corresponds to a dataset and training stages (Pretrain, SFT, and RLFT).
For RLFT, the batch size denotes the total number of samples per update (10 images × 6 generations).
Rows such as $\lambda_{\mathrm{PVB}}$ and $\lambda_{\mathrm{L2}}$ are specific to SFT stages only, described in Eq.~\ref{eq:sft}. 
A dash indicates that the parameter is not applicable.}
\begin{tabular}{l|ccc|ccc|cc}
\toprule
\multirow{2}{*}{\textbf{Hyperparameter}} 
& \multicolumn{3}{c|}{\textbf{MetalSet}} 
& \multicolumn{3}{c|}{\textbf{ViaSet}} 
& \multicolumn{2}{c}{\textbf{StdContact}} \\
\cmidrule(lr){2-4} \cmidrule(lr){5-7} \cmidrule(lr){8-9}
& Pretrain & SFT & RLFT 
& Pretrain & SFT & RLFT 
& SFT & RLFT \\
\midrule
Epochs & 50 & 25 & - & 10 & 1 & - & 100 & - \\
Batch size & 12 & 12 & 60 & 12 & 12 & 60 & 12 & 60 \\
Learning rate & 1e-4 & 2e-5 & 2e-6 & 1e-4 & 2e-5 & 2e-6 & 2e-5 & 2e-6 \\
$\lambda_{\mathrm{PVB}}$ & – & 0.248 & – & – & 2e-5 & – & 2e-5 & – \\
$\lambda_{\mathrm{L2}}$ & – & 0.002 & – & – & 2e-5 & – & 2e-5 & – \\
\midrule
$\lambda_{\mathrm{flow}}$ & \multicolumn{8}{c}{$1 - \lambda_{\mathrm{PVB}} - \lambda_{\mathrm{L2}}$} \\
\bottomrule
\end{tabular}

\label{tab:exp_settings}
\end{table}

\section{Supplementary Results}
\label{sec:supplementary_results}

\subsection{Noise Type and Level on ViaSet}
\begin{figure*}[!th]
  \centering
  \includegraphics[width=\textwidth]{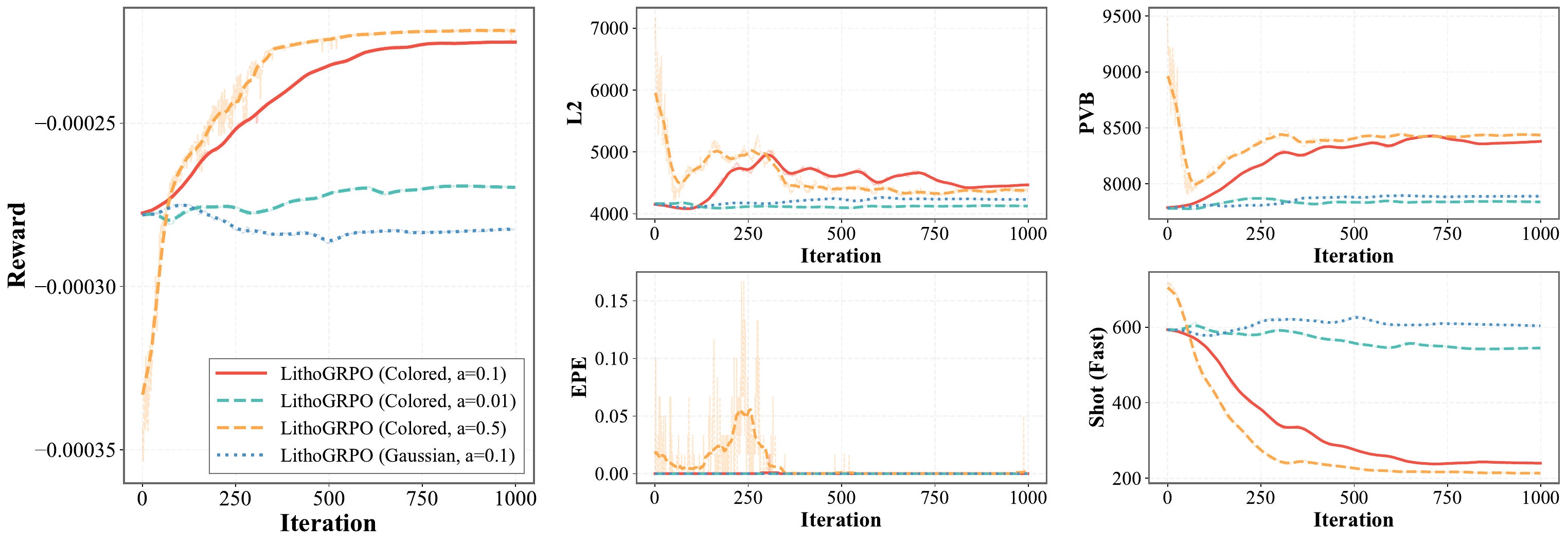}
  \caption{\textbf{Noise configurations in SDE (ViaSet).} 
We compare noise types (Colored vs. Gaussian) and noise levels ($a \in \{0.01, 0.1, 0.5\}$) for LithoGRPO. 
}
  \label{fig:ablation_viaset}
\end{figure*}

Similar to Fig.~\ref{fig:ablation_metalset}, we also report the training status with different noise types and levels in the RLFT stage. The results are shown in Fig.~\ref{fig:ablation_viaset}. 
Overall, the conclusions are consistent with the observations on the MetalSet experiments: the \textit{colored} noise with $a=0.1$ still demonstrates more stable and generally better performance. Although the case with $a=0.5$ yields a slightly higher reward in this particular setting, it fails to achieve optimal and stable results across all datasets and also exhibits slower convergence.

\subsection{Fast Shot Count Estimation}
\label{sec:fast_shot_estimation}

We employ a fast shot count estimator to accelerate the reward computation during RLFT. Here we analyze the accuracy and reliability of this estimator compared to the ground-truth shot count computed by the full algorithm.

\begin{figure}[!ht]
  \centering
  \includegraphics[width=\columnwidth]{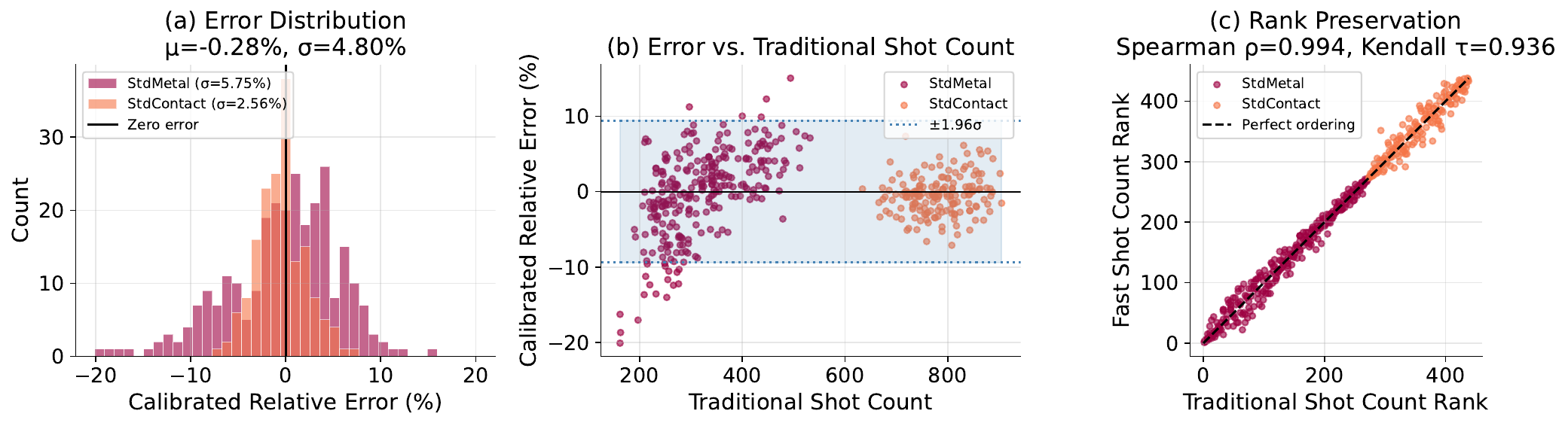}
  \caption{\textbf{Analysis of the fast shot count estimator.} 
  (a) Distribution of calibrated relative errors across StdMetal and StdContact benchmarks. 
  (b) Calibrated error as a function of traditional shot count, with 95\% confidence bounds shown. 
  (c) Rank preservation between the fast estimator and traditional shot counts, demonstrating high correlation (Spearman $\rho = 0.994$, Kendall $\tau = 0.936$).}
  \label{fig:error_and_rank_analysis}
\end{figure}

We evaluate the fast estimator on both the StdMetal and StdContact benchmarks generated by LithoGRPO (RLFT). 
We fit a linear regression between the fast estimator and the exact shot count. 
The residual errors exhibit a near-zero mean ($\mu = -0.28\%$) with a standard deviation of $\sigma = 4.80\%$, and 95\% of all errors fall within $\pm 9.54\%$. 
These results are summarized in Fig.~\ref{fig:error_and_rank_analysis}(a, b).

More importantly, for reinforcement learning, the ranking consistency of rewards matters more than their absolute accuracy, since policy gradient methods depend primarily on relative comparisons among samples within each batch. 
Furthermore, as the reward is normalized within each batch during training, any systematic bias or scale discrepancy in the estimator is effectively eliminated, leaving only the relative ordering to influence gradient updates.
As shown in Fig.~\ref{fig:error_and_rank_analysis}(c), the fast estimator achieves a Spearman correlation of $\rho = 0.994$ and a Kendall's $\tau = 0.936$, indicating near-perfect preservation of the relative ordering. 
This strong rank correlation ensures that the RL training signal remains reliable.

\subsection{Sensitivity Analysis of Metric Weights}

\begin{figure}[t]
\centering
\includegraphics[width=\linewidth]{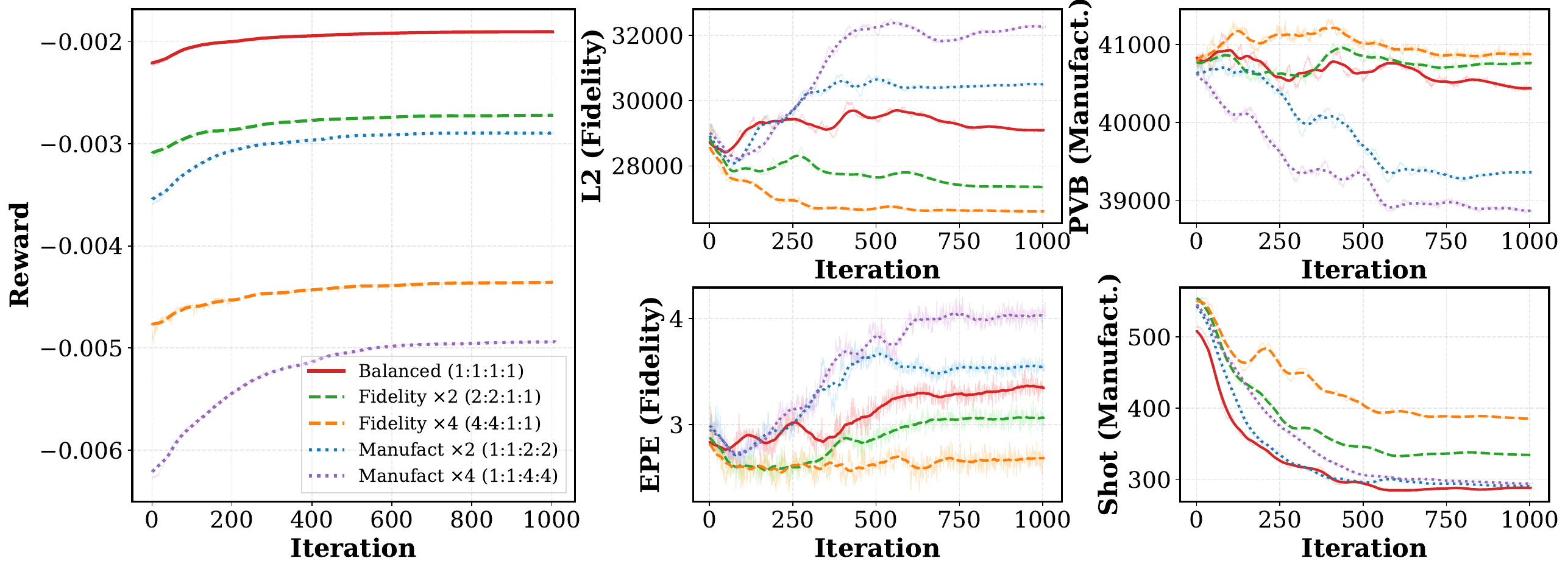}
\caption{\textbf{Sensitivity analysis of reward metric weights across different configurations on MetalSet.} Increasing fidelity weights (L2, EPE) improves fidelity metrics at the cost of manufacturability (PVB, Shot), and vice versa. Despite 4× weight variations, all configurations maintain stable training.}
\label{fig:weight_sensitivity}
\end{figure}

To evaluate the robustness of our reward weighting scheme, we conduct sensitivity analysis by varying the relative weights of fidelity metrics (L2, PVB) and manufacturability metrics (EPE, Shot), as shown in Fig.~\ref{fig:weight_sensitivity}. We evaluate five configurations: balanced (1:1:1:1), fidelity-emphasized (2:2:1:1 and 4:4:1:1), and manufacturability-emphasized (1:1:2:2 and 1:1:4:4).

The results reveal expected trade-off behaviors: increasing fidelity weights leads to improved L2 and EPE scores but degrades PVB and Shot performance, while emphasizing manufacturability weights yields the opposite effect. 
This trade-off scales proportionally with the weight magnitude: configurations with 4× weights show more pronounced metric shifts than those with 2× weights. 
Notably, despite these variations, all configurations exhibit stable optimization trajectories without training collapse or divergence.
This demonstrates that LithoGRPO is robust to weight perturbations, allowing practitioners to adjust weights according to application-specific priorities without risking training instability.

\subsection{Visualization on ViaSet}

\begin{figure*}[!th]
  \centering
    \includegraphics[width=\textwidth]{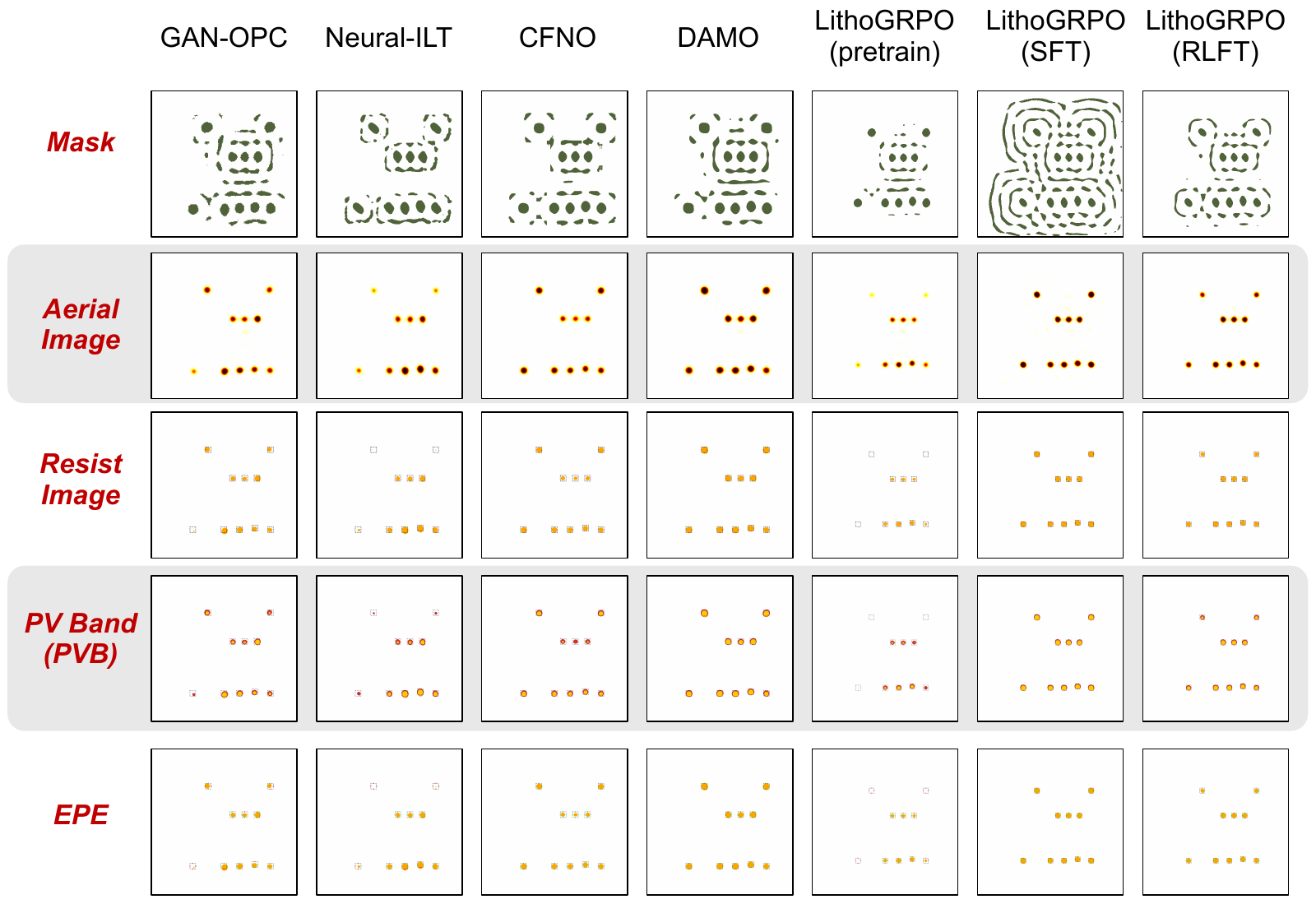}
    \label{fig:viaset_visualization}
\caption{\textbf{Visualization of ILT results on the ViaSet case.}}
  \label{fig:teaser_viaset}
\end{figure*}

We visualize a representative case from the ViaSet test set in comparison with the baselines and our LithoGRPO in Fig.~\ref{fig:teaser_viaset}. Our method shows competitive results.

\end{document}